\documentclass[11pt]{article}

\usepackage[preprint]{acl}

\usepackage{times}
\usepackage{latexsym}

\usepackage[T1]{fontenc}
\usepackage[utf8]{inputenc}

\usepackage{microtype}

\usepackage{inconsolata}

\usepackage{graphicx}

\usepackage[utf8]{inputenc} % allow utf-8 input
\usepackage[T1]{fontenc}    % use 8-bit T1 fonts
\usepackage{hyperref}       % hyperlinks
\usepackage{url}            % simple URL typesetting
\usepackage{booktabs}       % professional-quality tables
\usepackage{amsfonts}       % blackboard math symbols
\usepackage{graphicx}       % for \resizebox
\usepackage{nicefrac}       % compact symbols for 1/2, etc.
\usepackage{microtype}      % microtypography
\usepackage{xcolor}         % colors
\usepackage{enumitem}
\usepackage{amsmath}
\usepackage{pifont}
\usepackage{makecell}

\usepackage{xcolor}
\usepackage[most]{tcolorbox}
\usepackage{amsmath}     
\usepackage{amssymb}     
\usepackage{needspace}
\usepackage{afterpage}
\usepackage{wrapfig}
\usepackage{bm} 
\usepackage{multirow}
\usepackage{subcaption}
\usepackage{graphicx}
\usepackage{xcolor}
\usepackage[normalem]{ulem}
\usepackage{colortbl}
\usepackage{booktabs}
\usepackage{mathtools}
\definecolor{methodbg}{RGB}{245,245,245}

\usepackage[dvipsnames,table]{xcolor}
\definecolor{bestcolor}{HTML}{F4E3F7}      % lavender/pink tint (best)
\definecolor{secondcolor}{HTML}{E9EFFA}    % pale blue/gray tint (second best)

\definecolor{bestcolor}{HTML}{C199F5}    % 较深紫色 — 高亮“最佳方法” 
\definecolor{secondcolor}{HTML}{D4ECFF}
\definecolor{lightgray}{gray}{0.9}
\definecolor{darkred}{RGB}{150,0,0}
\definecolor{lightred}{RGB}{255,200,200}
\definecolor{lightblue}{RGB}{220,220,255}

\usepackage{amsthm}

\newtheorem{lemma}{Lemma}
\newtheorem{theorem}{Theorem}
\newtheorem{remark}{Remark}

\title{AutoL2S: Auto Long-Short Reasoning for Efficient LLMs}

\def\Algname{Auto Long-Short Reasoning}
\def\Algnameabbr{AutoL2S}

\author{
 \textbf{Feng Luo~\thanks{Equal Contribution.}\textsuperscript{1}},
 \textbf{Yu-Neng Chuang$^*$\textsuperscript{1}},
 \textbf{Guanchu Wang\textsuperscript{2}},
 \textbf{Hoang Anh Duy Le\textsuperscript{1}},
 \textbf{Shaochen Zhong\textsuperscript{1}},
 \\
 \textbf{Hongyi Liu\textsuperscript{1}},
 \textbf{Jiayi Yuan\textsuperscript{1}},
 \textbf{Yang Sui\textsuperscript{1}},
 \textbf{Vladimir Braverman\textsuperscript{1}},
 \textbf{Vipin Chaudhary\textsuperscript{4}},
 \textbf{Xia Hu\textsuperscript{1}}
\\
 \textsuperscript{1}Rice University,
 \textsuperscript{2}University of North Carolina at Charlotte,
 \textsuperscript{3}John Hopkins University
 \\
 \textsuperscript{4}Case Western Reserve University.
}

\begin{document}
\maketitle

\begin{abstract}
  Reasoning-capable large language models (LLMs) achieve strong performance on complex tasks but often exhibit overthinking after distillation, generating unnecessarily long chain-of-thought (CoT) reasoning even for simple inputs and incurring high inference cost. However, naively shortening reasoning length can degrade reasoning accuracy, as concise reasoning may be insufficient for certain inputs and lacks explicit supervision.
  We propose Auto Long-Short Reasoning (\Algnameabbr{}), a distillation framework that empowers non-reasoning LLMs to think thoroughly but only when necessary. 
  \Algnameabbr{} first learns a lightweight switching token with verified long-short CoTs to enable instance-wise long-short reasoning selection. Then it leverages long-short reasoning rollouts induced by switching tokens within a GRPO-style loss to improve reasoning efficiency while maintaining accuracy.
  Experiments demonstrate that \Algnameabbr{} effectively reduces reasoning length up to 71\% with minimal accuracy loss, yielding markedly better trade-off in token length and inference time while preserving accuracy. 
  % The code is available at \url{https://anonymous.4open.science/r/AutoL2S-A72E} 
\end{abstract}

% \gc{Guanchu: Need full-text unified: reasoning path, CoT reasoning path, CoT reasoning, CoT reasoning path, CoTs, so many terms for one concept. Keep 'reasoning' for defining the problem, and CoT reasoning path for defining the data (CoTs for abbreviation).}

\section{Introduction}

Reasoning distillation is an effective approach for transferring complex reasoning abilities from strong teacher large language models (LLMs) to non-reasoning capable student LLMs~\cite{guo2025deepseek,bespoke_stratos,muennighoff2025s1,ye2025limo}, but it often introduces a critical, inefficient overthinking issue~\cite{sui2025stop}. Distilled models tend to generate excessively long CoT reasoning paths even for inputs that admit concise solutions, resulting in substantial increases in decoding time, memory usage, and deployment cost~\cite{chen2024not}.
This behavior arises because distillation typically trains student models to imitate full long-form reasoning paths in order to preserve accuracy, implicitly treating long reasoning as uniformly necessary across instances. As a result, distilled models lack signals indicating when shorter reasoning would suffice.
Existing approaches mitigate overthinking through manual post-distillation control of reasoning modes (e.g., prompting users to select short or long reasoning)~\cite{qwen2.5,claude}, or by learning reasoning-mode selection via special tokens or reinforcement learning guided by outcome-based rewards~\cite{luo2025o1,ma2025cot,fang2025thinkless}. However, these methods rely on strong reference behaviors and often suffer from accuracy degradation when reasoning is aggressively shortened.

The challenges lie in the nature of the trade-off between reasoning length and accuracy.
First, short reasoning may lead to performance degradation compared to long ones, particularly on inputs that require multi-step inference or error correction. Without reliable signals indicating when compression is safe, enforcing shorter reasoning risks discarding intermediate steps that are critical to correctness, resulting in inferior and inconsistent behaviors across inputs~\cite{luo2025o1, fang2025thinkless}.
Second, reasoning length is difficult to regulate while preserving accuracy. For many inputs, multiple reasoning paths of varying lengths can lead to correct answers, and whether concise reasoning is sufficient is often only observable through outcome correctness~\cite{ma2502cot,zhang2025lessons,liu2024can,yu2024distilling}.
These challenges suggest that effective CoT compression cannot be treated as uniform truncation, but must instead be framed as an instance-wise decision problem that dynamically balances correctness and token-generation cost.
We ask: \textbf{\textit{How can reasoning distillation allow reasoning length to vary across instances while maintaining correctness?}}

To address these challenges, we propose Auto Long-Short Reasoning (\Algnameabbr{}), a distillation framework that enables non-reasoning LLMs to reason thoroughly only when necessary. \Algnameabbr{} pairs long teacher reasoning with verified short reasoning paths and explicitly supervises a lightweight switching token (\texttt{<EASY>}) during distillation. The \texttt{<EASY>} token is generated as part of the reasoning sequence and induces instance-wise selection between long and short CoT reasoning paths. Building on this supervision, \Algnameabbr{} further leverages the induced long-short reasoning rollouts by \texttt{<EASY>} token to fine-tune the model with a GRPO-style loss. This stage encourages correctness of CoT reasoning paths while implicitly favoring shorter reasoning when sufficient, allowing the model to internalize instance-wise trade-offs between reasoning sufficiency and efficiency.

Across multiple reasoning benchmarks, we demonstrate that \Algnameabbr{} substantially reduces reasoning length while maintaining accuracy, achieving up to a 71\% reduction in reasoning length with minimal loss in accuracy. The contributions are listed as follows:

\vspace{-0.2cm}
\begin{itemize}[leftmargin=10pt]
    \itemsep=-1pt
    \item \textbf{Auto Long-Short Reasoning.} We propose \Algnameabbr{}, a distillation framework that pairs long and verified short reasoning paths for long-short reasoning selection, and further improves efficiency via GRPO-style fine-tuning under supervision-induced long-short rollouts.
    \item \textbf{Implicit Control via Joint Generation.} We introduce a lightweight \texttt{<EASY>} switching token that is jointly generated before the actual rollout, allowing long-short mode selection to be learned implicitly without explicit length constraints.
    \item \textbf{Reasoning Evaluation.} \Algnameabbr{} achieves substantial efficiency gains while maintaining accuracy across multiple reasoning benchmarks.  
\end{itemize}

\section{Preliminary}

In this section, we first formally define the Auto Long-Short reasoning problem. We then illustrate the challenges in controlling the reasoning length of distilled large reasoning-capable LLMs.

\begin{figure*}
    \centering
    \includegraphics[width=1\linewidth]{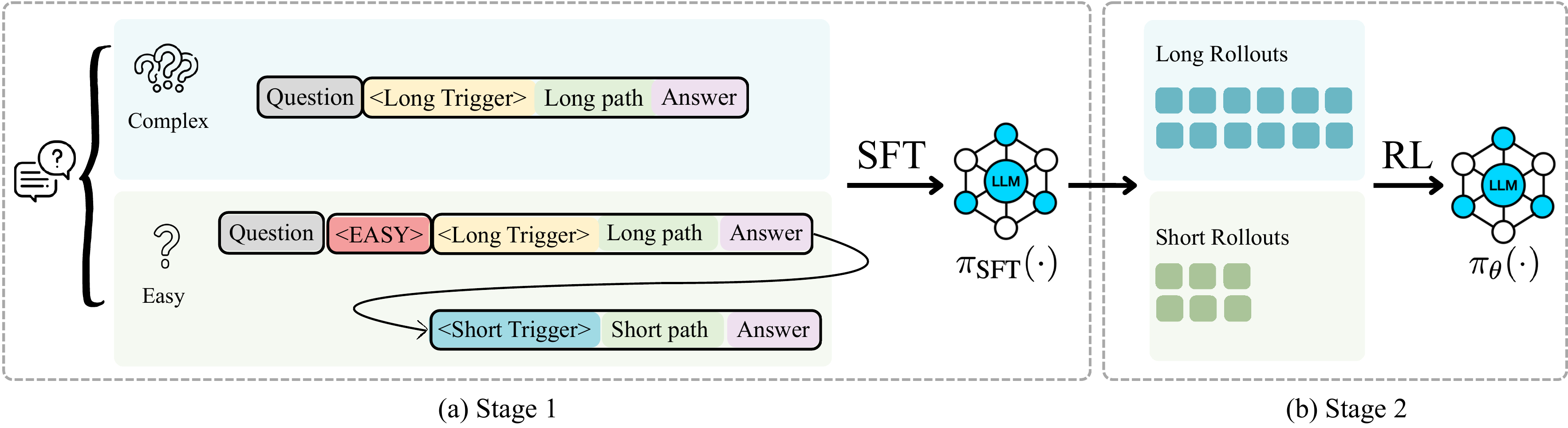}
    \vspace{-.7cm}
    \caption{Training pipeline of \Algnameabbr{} in two stages. (a) Supervised with paired long and short CoT reasoning paths with \texttt{<EASY>} token. (b) \texttt{<EASY>} token is used to induce long-short reasoning rollout, enabling the model to generate short or long reasoning paths in an instance-dependent manner.}
    \label{fig:autol2s-data}
    \vspace{-3mm}
\end{figure*}

\subsection{Problem Definition}
We aim to develop reasoning models $\pi(\cdot \mid \theta)$ with trainable parameters $\theta$ that complete tasks correctly while using reasoning paths as short as possible.
The objective is to train $\pi(\cdot \mid \theta)$ in $\mathcal{D}$ to learn a policy that selects minimal sufficient reasoning for each input.
We expect the outputs of $\pi(\cdot \mid \theta_{\mathcal{D}})$ to be sufficiently short while maintaining reasoning accuracy.
This reduction in output length translates directly to fewer generated tokens and thus faster inference. To this end, we propose the \textit{Auto Long-Short Reasoning}~(\Algnameabbr{}) framework to enable efficient LLM reasoning through joint utilization of valid long and short CoT reasoning paths.
We emphasize that reasoning length does not admit a unique ground-truth label: for many inputs, multiple reasoning paths of different lengths can lead to correct answers. \Algnameabbr{} aims to identify an effective trade-off between reasoning accuracy and generation efficiency. Therefore, \Algnameabbr{} is evaluated based on outcome preservation under reasoning length with accuracy preservation.

\subsection{Challenges of Length Controlling} 

Balancing brevity and completeness in reasoning remains challenging. Aggressive compression of reasoning paths can omit essential intermediate steps, leading to degraded performance on complex inputs, while the absence of supervision signals for the minimally sufficient reasoning trace makes it difficult for models to determine when concise reasoning is appropriate.
Recent SFT-based approaches mitigate overthinking by curating datasets with variable-length or information-dense reasoning traces and fine-tuning models to produce shorter reasoning~\cite{ma2025cot, xia2025tokenskip}. However, these methods primarily encourage global compression and do not provide an explicit mechanism for instance-wise reasoning-length selection, often resulting in over-compression on inputs that require extended reasoning. Related work~\cite{fang2025thinkless} explores reasoning-mode selection using switching tokens and reinforcement learning, but such approaches rely on outcome-based rewards rather than direct supervision of correctness-preserving brevity, and lack explicit signals indicating when shorter reasoning suffices. As a result, models do not receive explicit supervision, distinguishing cases where concise reasoning is sufficient from those that require extended reasoning, relying instead on outcome-based optimization, which can make instance-wise reasoning-length adaptation less reliable. These limitations suggest that effective reasoning compression cannot rely solely on uniform supervision or reward-driven mode selection, but instead requires structured supervision that enables reasoning length to vary across instances.

\section{\Algname{}}
We systematically introduce the \Algnameabbr{} framework. \Algnameabbr{} aims to distill reasoning capabilities from reasoning-capable LLMs, allowing the model to learn effective reasoning patterns while reducing the length of reasoning paths required to arrive at correct reasoning answers. To achieve our goal, we propose a two-stage training pipeline: (1) supervised fine-tuning for adaptive reasoning rollout selection, and (2) leveraging GRPO-style optimization with supervision-induced rollouts.

\subsection{SFT Stage of \Algnameabbr{}}

\Algnameabbr{} constructs a diverse reasoning dataset containing both long and short CoT reasoning paths based on prediction correctness. The construction pipeline is illustrated in Figure~\ref{fig:autol2s-data}. Long CoT reasoning paths are provided for all questions to capture complete reasoning, while short CoT reasoning paths are preferred whenever they still yield correct answers, offering more efficient representations. \Algnameabbr{} trains LLMs to learn both long and short reasoning paths and to identify EASY questions, enabling efficient reasoning when appropriate.

\vspace{0.1cm}
\noindent\textbf{Constructing Long CoT Reasoning Paths.} We use Bespoke-Stratos-17k~\cite{bespoke_stratos} as the source of questions and employ a strong reasoning-capable LLM as teacher to generate long CoT reasoning paths together with final answers, forming the base long-CoT dataset. For an input $X$ with ground-truth answer $y^\star$, we treat $L$ as an effective long reasoning path if it yields the correct answer, without requiring token-level semantic optimality.

\vspace{0.1cm}
\noindent\textbf{Constructing Short CoT Reasoning Paths.} 
To avoid uniformly enforcing long reasoning when concise reasoning suffices, we generate a short reasoning path $S$ such that $|S| \ll |L|$. Specifically, we employ a short CoT teacher to generate candidate short reasoning paths $\{S_j\}_{j=1}^{k}$ using rejection sampling with $k$ trials. 
Among these candidates that yield the correct answer, we select the shortest path $S = \arg\min_{\{S_j \mid \hat{y}(S_j) = y\}} |S_j|$ as the effective short CoT reasoning path.
This procedure yields concise reasoning traces that preserve correctness while minimizing reasoning length.

\vspace{0.1cm}
\noindent\textbf{SFT Training Strategy.}
\Algnameabbr{} follows Figure~\ref{fig:autol2s-data} to construct $\mathcal{D}$, where special tokens \texttt{<EASY>}, \texttt{<Long Trigger>}, \texttt{<Short Trigger>}, and \texttt{<Answer Trigger>} are used to hook the questions, long-short reasoning, and final answers. For inputs admitting both valid long and short CoT reasoning paths $(L, S)$, we annotate the question with the \texttt{<EASY>} token; for inputs where no such valid short CoT exists, we retain the original long reasoning $L$ and omit the \texttt{<EASY>} token.
Formally, \Algnameabbr{} adopts the constructed dataset $\mathcal{D} =\{(x_i,\mathcal{R}_i,y_i^\star)\}_{i=1}^N$, where each reasoning path $\mathcal{R}_i\in\{\{L_i\},\{\texttt{<EASY>},L_i,S_i\}\}$, and trains the model $\pi(\cdot\mid\theta)$ by minimizing the next-token prediction loss function given as follows:
\begin{equation}
    \label{eq:train}
    \notag
    \mathcal{L}_{\text{SFT}} = -\mathbb{E}_{(x_i,\mathcal{R}_i,y_i^\star) \sim \mathcal{D}} \big[\log \pi(r_t ~|~ r_{<t}, x_i, \theta) \big],
\end{equation}
where $r_{<i}$ denotes the prefix tokens in $\mathcal{R}_t$ that precede position $t$. We denote the SFT-trained model as $\pi_{\text{SFT}}(\cdot)$.

\subsection{\Algnameabbr{}: Long-Short Joint Rollouts}
In this section, we correct residual length-accuracy misalignment induced by the SFT stage using positive and negative signals from long-short rollouts.

\subsubsection{Long-short Rollout Generation}
\begin{figure}
\centering
\begin{subfigure}[t]{0.45\textwidth}
    \includegraphics[width=\textwidth]{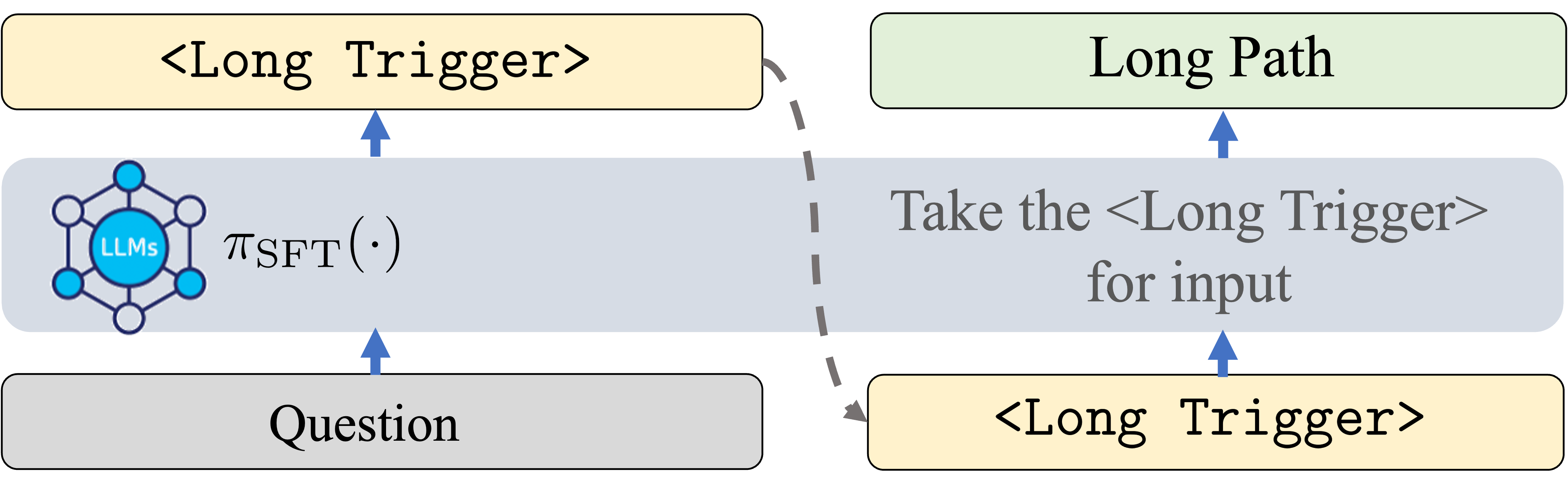}
    \vspace{-6mm}
    \caption{Generating the long reasoning rollout}
    \vspace{0.2cm}
\end{subfigure}
\hfill
\begin{subfigure}[t]{0.45\textwidth}
    \includegraphics[width=\textwidth]{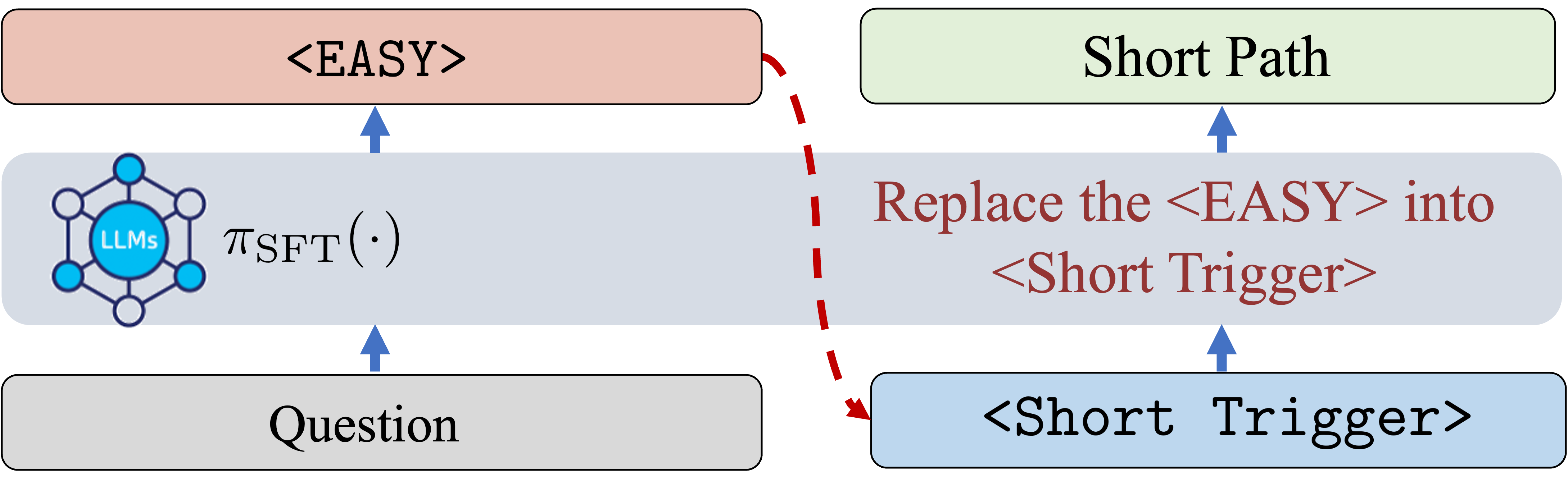}
    \vspace{-3mm}
    \caption{Generating the short reasoning rollout}
\end{subfigure}
\vspace{-0.1cm}
\caption{\label{fig:inference} The SFT model $\pi_{\text{SFT}}(\cdot)$ generates (a) a long reasoning rollout when \texttt{<Long Trigger>} yields; or (b) a short reasoning rollout conditioned on \texttt{<EASY>} token. The long-short selection is determined by $\pi_{\text{SFT}}(\cdot)$.}
\vspace{-4mm}
\end{figure}

During the inference stage, the SFT model $\pi_{\text{SFT}}(\cdot)$ is able to determine whether short CoT reasoning is sufficient to solve questions, enabling adaptive length for rollout generation.
Specifically, as illustrated in Figure~\ref{fig:inference}, $\pi_{\text{SFT}}(\cdot)$ begins generation by producing either a \texttt{<Long Trigger>} or a \texttt{<EASY>} token, which determines the subsequent CoT generation.
If the first generated token is a \texttt{<Long Trigger>} token (as shown in Figure~\ref{fig:inference}(a)), it indicates that the question requires a long reasoning path, and then the model proceeds with standard autoregressive generation to complete the long reasoning and produce the final answer. In contrast, if $\pi_{\text{SFT}}(\cdot)$ initially generates an \texttt{<EASY>} token (as shown in Figure~\ref{fig:inference}(b)), which suggests the question is solvable with a short CoT. Then, we leverage constraint decoding to directly inject \texttt{<Short Trigger>} token to enforce the model generating short reasoning paths.

\subsubsection{Joint Rollouts Training}

Leveraging automatic long-short selection with \texttt{<EASY>} token, \Algnameabbr{} proposes a GRPO-style framework to further improve reasoning efficiency while maintaining accuracy. Specifically, it leverages the long-short CoT rollouts generated by the SFT model $\pi_{\text{SFT}}(\cdot)$, and proposes a GRPO-style objective function that maximizes the model accuracy while keeping the output length distribution close to the SFT model, preserving the foundation capability for long-short selection. The objective function is formally defined by:
\begin{align}
\label{eq:grpo-loss}
\notag
    \mathcal{J}_{\text{AutoL2S}}(\theta) = &\mathbb{E}_{\substack{z \sim \mathcal{D} \\ r \sim \pi_{\text{SFT}}}} \Big[ \text{min} \big( w(\theta) A(z, r), \\
    &\text{clip}(w(\theta), 1 - \varepsilon, 1 + \varepsilon) A(z, r) \big) \Big], \nonumber
\end{align}
where $w(\theta) = \frac{\pi_{\theta}(r | z)}{\pi_{\text{SFT}}(r | z)}$ indicates the ratio between the trainable model $\pi_{\theta}(\cdot)$ and SFT model $\pi_{\text{SFT}}(\cdot)$; $A(z, r) \!=\! \mathbf{U}(z, r) \!-\! \mathbb{E}_{r_i \sim \pi_{\text{SFT}}}[\mathbf{U}(z, r_i)]$ denotes the advantage of a rollout $r$, where $\mathbf{U}(z, r)$ takes 1 or 0, corresponding to whether $r$ derives correct answers to a question $z$ or not; and $\text{clip}(\cdot, 1 \!-\! \varepsilon, 1 \!+\! \varepsilon)$ is a clipping function with $0 \leq \varepsilon \leq 1$. The $\text{clip}(\cdot)$ operator constrains the output distribution to remain close to the SFT model.

The intuition behind \Algnameabbr{} is a combination of exploitation and exploration.
\textit{Exploitation:} $\pi_{\theta}(\cdot)$ is initialized from $\pi_{\text{SFT}}(\cdot)$, ensuring initial capability for long-short CoT selection for adaptive rollout generation.
\textit{Exploration:} The rollout advantage $A(z, r)$ serves as a correctness signal: $A(z, r) \!\textgreater\! 0$ if $r$ yields correct answers, and $A(z, r) \!\textless\! 0$ otherwise. It encourages the model to explore correct solutions by reinforcing successful outcomes and discouraging incorrect ones with dynamic length of rollouts, thereby strengthening the adaptive reasoning behaviors learned after the SFT stage.

\subsection{Theoretical Interpretation of \Algnameabbr{}}

In this section, we provide a theoretical interpretation of \Algnameabbr{} to clarify the mechanisms underlying its training procedure. Lemma~\ref{lemma:1} in Appendix~\ref{apdx:thm} offers an information-theoretic perspective, suggesting that conditioning short CoT reasoning paths on long reasoning can reduce learning uncertainty by supplying additional contextual information. This intuition naturally extends to the \Algnameabbr{} training setting, where optimization is performed via cross-entropy (equivalently, perplexity): concatenating long CoT paths with short ones effectively enriches the supervisory signal available for learning better short reasoning behaviors.

% Rejection sampling provides a principled way to operate this idea. By filtering candidate short CoT reasoning paths for correctness, rejection sampling controls how much auxiliary information from long reasoning is retained, encouraging short paths that are both concise and aligned with their long counterparts. In this sense, long reasoning acts as auxiliary context that stabilizes short-path learning without requiring explicit length constraints.

At the same time, reasoning length does not admit a unique ground-truth label. For many inputs, multiple reasoning paths of different lengths can yield correct answers, making long-short reasoning selection inherently instance-dependent. This motivates viewing reasoning-length choice as a trade-off between generation cost and uncertainty. We formalize this intuition in Theorem~\ref{apex:thm:2} in Appendix~\ref{apdx:thm}, which frames long-short reasoning selection as a risk-cost trade-off: shorter reasoning reduces token cost but may increase uncertainty, while longer reasoning provides redundancy at higher computational expense.

From this unified perspective, paired long-short CoT supplies auxiliary information that reduces uncertainty when learning concise reasoning behaviors, thereby stabilizing short-path rollout generation while preserving correctness. This analysis is not used to derive the training objective of \Algnameabbr{}, nor does it claim optimality or theoretical guarantees. Rather, it serves as a conceptual lens for understanding why structured long-short supervision supports adaptive reasoning compression, consistent with the empirical results in Section~\ref{sec:experiments}.

\begin{table*}[t]
\small
\caption{Accuracy (Acc) and Token Length (Len) across six reasoning benchmarks. 
Values in parentheses denote the accuracy improvement and token reduction relative to the Bespoke-Stratos-3B/7B model. ``AutoL2S-SFT" defines AutoL2S with only SFT Stage, and ``w/o RJ" defines AutoL2S without rejection sampling. Purple and blue cells highlight the best and second-best values, respectively.}
\centering
\resizebox{\textwidth}{!}{%
\begin{tabular}{l
c c
c c
c c
c c
c c
c c
c c}
\toprule
\textbf{}
& \multicolumn{2}{c|}{\textbf{Average}}
& \multicolumn{2}{c|}{\textbf{MATH500}}
& \multicolumn{2}{c|}{\textbf{GPQA}}
& \multicolumn{2}{c|}{\textbf{GSM8K}}
& \multicolumn{2}{c|}{\textbf{Olympiad}}
& \multicolumn{2}{c|}{\textbf{AIME}}
& \multicolumn{2}{c}{\textbf{MMLU-Pro}} \\
\cmidrule(lr){2-3}
\cmidrule(lr){4-5}
\cmidrule(lr){6-7}
\cmidrule(lr){8-9}
\cmidrule(lr){10-11}
\cmidrule(lr){12-13}
\cmidrule(lr){14-15}
& Acc & Len
& Acc & Len
& Acc & Len
& Acc & Len
& Acc & Len
& Acc & Len
& Acc & Len\\
\midrule

% ===================== 3B RESULTS (Updated Colors) =====================
\multicolumn{15}{c}{\textit{Llama-3.2-3B-Instruct}} \\
\midrule

Llama-3.2-3B-Instruct
& 0.357 & 1015
& 0.404 & 740
& 0.293 & 498
& 0.729 & 203
& 0.147 & 2117
& 0.067 & 2053
& 0.500 & 477 \\
\addlinespace[2pt]

Bespoke-Stratos-3B
& 0.413 & 10219
& \cellcolor{bestcolor}{\textbf{0.574}} & 10148
& 0.273 & 8888
& 0.822 & 1387
& \cellcolor{bestcolor}{\textbf{0.246}} & 15635
& 0.033 & 21341
& \cellcolor{bestcolor}{\textbf{0.529}} & 3912 \\
\addlinespace[2pt]

CoT-Valve
& 0.363 & 11941
& 0.478 & 10890
& 0.283 & 9634
& 0.773 & 2238
& 0.154 & 18634
& 0.033 & 26059
& 0.457 & 4191 \\
& (-0.050) & (+16.9\%)
& (-0.096) & (+7.3\%)
& (+0.010) & (+8.4\%)
& (-0.049) & (+61.4\%)
& (-0.092) & (+19.2\%)
& (+0.000) & (+22.1\%)
& (-0.072) & (+7.1\%) \\
\addlinespace[2pt]

O1-pruner
& 0.402 & 5995
& 0.562 & 5295
& 0.308 & 5394
& 0.816 & \cellcolor{secondcolor}{\textbf{860}}
& \cellcolor{secondcolor}{\textbf{0.236}} & \cellcolor{secondcolor}{\textbf{8622}}
& 0.033 & \cellcolor{secondcolor}{\textbf{12074}}
& 0.457 & 3724 \\
& (-0.011) & (-41.3\%)
& (-0.012) & (-47.8\%)
& (+0.035) & (-39.3\%)
& (-0.006) & \cellcolor{secondcolor}{\textbf{(-38.0\%)}}
& \cellcolor{secondcolor}{\textbf{(-0.010)}} & \cellcolor{secondcolor}{\textbf{(-44.9\%)}}
& (+0.000) & \cellcolor{secondcolor}{\textbf{(-43.4\%)}}
& (-0.072) & (-4.8\%) \\
\addlinespace[2pt]

DPO
& 0.399 & 7354
& \cellcolor{bestcolor}{\textbf{0.574}} & 5363
& 0.283 & 6740
& \cellcolor{bestcolor}{\textbf{0.832}} & 911
& 0.227 & 10441
& 0.033 & 17456
& 0.443 & 3215 \\
& (-0.014) & (-28.0\%)
& \cellcolor{bestcolor}{\textbf{(+0.000)}} & (-47.2\%)
& (+0.010) & (-24.2\%)
& \cellcolor{bestcolor}{\textbf{(+0.010)}} & (-34.3\%)
& (-0.019) & (-33.2\%)
& (+0.000) & (-18.2\%)
& (-0.086) & (-17.8\%) \\
\addlinespace[2pt]

TokenSkip
& 0.379 & 10579
& 0.512 & 10327
& 0.258 & 9438
& 0.801 & 2238
& 0.191 & 15853
& 0.000 & 21122
& \cellcolor{secondcolor}{\textbf{0.514}} & 4496 \\
& (-0.033) & (+3.5\%)
& (-0.062) & (+1.8\%)
& (-0.015) & (+6.2\%)
& (-0.021) & (+61.4\%)
& (-0.055) & (+1.4\%)
& (-0.033) & (-1.0\%)
& \cellcolor{secondcolor}{\textbf{(-0.015)}} & (+14.9\%) \\
\addlinespace[2pt]

% ===================== AutoL2S =====================
\rowcolor{methodbg}
\textbf{AutoL2S-SFT (w/o RJ)}
& \cellcolor{bestcolor}{\textbf{0.418}} & 8280
& 0.552 & 5990
& \cellcolor{bestcolor}{\textbf{0.389}} & 7520
& 0.823 & 1166
& 0.206 & 12941
& \cellcolor{secondcolor}{\textbf{0.067}} & 19158
& 0.471 & 2906 \\
\rowcolor{methodbg}
& \cellcolor{bestcolor}{\textbf{(+0.005)}} & (-19.0\%)
& (-0.022) & (-41.0\%)
& \cellcolor{bestcolor}{\textbf{(+0.116)}} & (-15.4\%)
& (+0.001) & (-15.9\%)
& (-0.040) & (-17.2\%)
& \cellcolor{secondcolor}{\textbf{(+0.034)}} & (-10.2\%)
& (-0.058) & (-25.7\%) \\
\addlinespace[2pt]

\rowcolor{methodbg}
\textbf{AutoL2S (w/o RJ)}
& 0.410 & \cellcolor{secondcolor}{\textbf{5954}}
& 0.564 & 6508
& 0.359 & \cellcolor{secondcolor}{\textbf{4569}}
& 0.815 & 964
& 0.230 & 9269
& 0.033 & 12494
& 0.457 & \cellcolor{secondcolor}{\textbf{1919}} \\
\rowcolor{methodbg}
& (-0.003) & \cellcolor{secondcolor}{\textbf{(-41.7\%)}}
& (-0.010) & (-35.9\%)
& (+0.086) & \cellcolor{secondcolor}{\textbf{(-48.6\%)}}
& (-0.007) & (-30.5\%)
& (-0.016) & (-40.7\%)
& (+0.000) & (-41.5\%)
& (-0.072) & \cellcolor{secondcolor}{\textbf{(-50.9\%)}} \\
\addlinespace[2pt]

\rowcolor{methodbg}
\textbf{AutoL2S-SFT}
& 0.389 & 6677
& 0.546 & \cellcolor{secondcolor}{\textbf{4181}}
& \cellcolor{secondcolor}{\textbf{0.369}} & 6165
& 0.800 & 1021
& 0.218 & 10706
& 0.000 & 14775
& 0.400 & 3211 \\
\rowcolor{methodbg}
& (-0.024) & (-34.7\%)
& (-0.028) & \cellcolor{secondcolor}{\textbf{(-58.8\%)}}
& \cellcolor{secondcolor}{\textbf{(+0.096)}} & (-30.6\%)
& (-0.022) & (-26.4\%)
& (-0.028) & (-31.5\%)
& (-0.033) & (-30.8\%)
& (-0.129) & (-17.9\%) \\
\addlinespace[2pt]

\rowcolor{methodbg}
\textbf{AutoL2S}
& \cellcolor{secondcolor}{\textbf{0.415}} & \cellcolor{bestcolor}{\textbf{4803}}
& 0.520 & \cellcolor{bestcolor}{\textbf{4116}}
& 0.273 & \cellcolor{bestcolor}{\textbf{3679}}
& \cellcolor{secondcolor}{\textbf{0.826}} & \cellcolor{bestcolor}{\textbf{819}}
& 0.193 & \cellcolor{bestcolor}{\textbf{7199}}
& \cellcolor{bestcolor}{\textbf{0.167}} & \cellcolor{bestcolor}{\textbf{11538}}
& \cellcolor{secondcolor}{\textbf{0.514}} & \cellcolor{bestcolor}{\textbf{1469}} \\
\rowcolor{methodbg}
& \cellcolor{secondcolor}{\textbf{(+0.002)}} & \cellcolor{bestcolor}{\textbf{(-53.0\%)}}
& (-0.054) & \cellcolor{bestcolor}{\textbf{(-59.4\%)}}
& (+0.000) & \cellcolor{bestcolor}{\textbf{(-58.6\%)}}
& \cellcolor{secondcolor}{\textbf{(+0.004)}} & \cellcolor{bestcolor}{\textbf{(-41.0\%)}}
& (-0.053) & \cellcolor{bestcolor}{\textbf{(-54.0\%)}}
& \cellcolor{bestcolor}{\textbf{(+0.134)}} & \cellcolor{bestcolor}{\textbf{(-45.9\%)}}
& \cellcolor{secondcolor}{\textbf{(-0.015)}} & \cellcolor{bestcolor}{\textbf{(-62.4\%)}} \\
\addlinespace[2pt]

\midrule\midrule
\multicolumn{15}{c}{\textit{Qwen2.5-7B-Instruct}} \\
\midrule

Qwen2.5-7B-Instruct
& 0.520 & 529
& 0.748 & 556
& 0.308 & 27
& 0.902 & 260
& 0.384 & 896
& 0.133 & 1014
& 0.643 & 423 \\
\addlinespace[2pt]

Bespoke-Stratos-7B
& 0.590 & 7430
& 0.824 & 5383
& 0.359 & 6049
& 0.926 & 1321
& 0.444 & 11322
& 0.200 & 18513
& \cellcolor{bestcolor}{\textbf{0.786}} & 1989 \\
\addlinespace[2pt]

CoT-Valve
& 0.543 & 5942
& 0.730 & 4483
& 0.369 & 4930
& 0.898 & 928
& 0.378 & 8647
& 0.167 & 14304
& 0.714 & 2362 \\
& (-0.047) & (-20.0\%)
& (-0.094) & (-16.7\%)
& (+0.010) & (-18.5\%)
& (-0.028) & (-29.7\%)
& (-0.066) & (-23.6\%)
& (-0.033) & (-22.7\%)
& (-0.072) & (+18.8\%) \\
\addlinespace[2pt]

O1-pruner
& 0.581 & 6773
& \cellcolor{bestcolor}{\textbf{0.832}} & 5104
& 0.399 & 5312
& \cellcolor{bestcolor}{\textbf{0.936}} & 1065
& 0.433 & 9586
& 0.200 & 17655
& 0.686 & 1916 \\
& (-0.009) & (-8.8\%)
& \cellcolor{bestcolor}{\textbf{(+0.008)}} & (-5.2\%)
& (+0.040) & (-12.2\%)
& \cellcolor{bestcolor}{\textbf{(+0.010)}} & (-19.4\%)
& (-0.011) & (-15.3\%)
& (+0.000) & (-4.6\%)
& (-0.100) & (-3.7\%) \\
\addlinespace[2pt]

DPO-Bespoke
& \cellcolor{secondcolor}{\textbf{0.593}} & 6073
& 0.806 & 3688
& 0.374 & 5961
& 0.920 & 1576
& \cellcolor{secondcolor}{\textbf{0.447}} & 7364
& \cellcolor{bestcolor}{\textbf{0.267}} & 15991
& 0.743 & 1858 \\
& \cellcolor{secondcolor}{\textbf{(+0.003)}} & (-18.3\%)
& (-0.018) & (-31.5\%)
& (+0.015) & (-1.5\%)
& (-0.006) & (+19.3\%)
& \cellcolor{secondcolor}{\textbf{(+0.003)}} & (-35.0\%)
& \cellcolor{bestcolor}{\textbf{(+0.067)}} & (-13.6\%)
& (-0.043) & (-6.6\%) \\
\addlinespace[2pt]

TokenSkip
& 0.565 & 7960
& \cellcolor{secondcolor}{\textbf{0.826}} & 5335
& \cellcolor{bestcolor}{\textbf{0.434}} & 5508
& 0.918 & 1165
& \cellcolor{secondcolor}{\textbf{0.447}} & 10947
& 0.067 & 22750
& 0.700 & 2054 \\
& (-0.024) & (+7.1\%)
& \cellcolor{secondcolor}{\textbf{(+0.002)}} & (-0.9\%)
& \cellcolor{bestcolor}{\textbf{(+0.075)}} & (-9.0\%)
& (-0.008) & (-11.8\%)
& \cellcolor{secondcolor}{\textbf{(+0.003)}} & (-3.3\%)
& (-0.133) & (+22.9\%)
& (-0.086) & (+3.3\%) \\
\addlinespace[2pt]

AlphaOne
& 0.519 & 4441
& 0.732 & 3867
& 0.313 & 6278
& 0.907 & 1943
& 0.356 & 5252
& 0.133 & 7162
& 0.671 & 2146 \\
& (-0.071) & (-40.2\%)
& (-0.092) & (-28.2\%)
& (-0.046) & (+3.8\%)
& (-0.019) & (+47.1\%)
& (-0.088) & (-53.6\%)
& (-0.067) & (-61.3\%)
& (-0.115) & (+7.9\%) \\
\addlinespace[2pt]

% ===================== AutoL2S =====================
\rowcolor{methodbg}
\textbf{AutoL2S-SFT (w/o RJ)}
& \cellcolor{bestcolor}{\textbf{0.600}} & 6314
& 0.800 & 3468
& \cellcolor{bestcolor}{\textbf{0.434}} & 4777
& \cellcolor{secondcolor}{\textbf{0.934}} & 735
& \cellcolor{bestcolor}{\textbf{0.470}} & 9068
& \cellcolor{secondcolor}{\textbf{0.233}} & 18332
& 0.729 & 1504 \\
\rowcolor{methodbg}
& \cellcolor{bestcolor}{\textbf{(+0.010)}} & (-15.0\%)
& (-0.024) & (-35.6\%)
& \cellcolor{bestcolor}{\textbf{(+0.075)}} & (-21.0\%)
& \cellcolor{secondcolor}{\textbf{(+0.008)}} & (-44.4\%)
& \cellcolor{bestcolor}{\textbf{(+0.026)}} & (-19.9\%)
& \cellcolor{secondcolor}{\textbf{(+0.033)}} & (-1.0\%)
& (-0.057) & (-24.4\%) \\
\addlinespace[2pt]

\rowcolor{methodbg}
\textbf{AutoL2S (w/o RJ)}
& 0.561 & \cellcolor{secondcolor}{\textbf{2299}}
& 0.798 & \cellcolor{secondcolor}{\textbf{1601}}
& 0.414 & \cellcolor{bestcolor}{\textbf{2666}}
& 0.912 & 707
& 0.439 & \cellcolor{secondcolor}{\textbf{3088}}
& 0.100 & \cellcolor{secondcolor}{\textbf{4638}}
& 0.700 & 1091 \\
\rowcolor{methodbg}
& (-0.029) & \cellcolor{secondcolor}{\textbf{(-69.1\%)}}
& (-0.026) & \cellcolor{secondcolor}{\textbf{(-70.3\%)}}
& (+0.055) & \cellcolor{bestcolor}{\textbf{(-55.9\%)}}
& (-0.014) & (-46.5\%)
& (-0.005) & \cellcolor{secondcolor}{\textbf{(-72.7\%)}}
& (-0.100) & \cellcolor{secondcolor}{\textbf{(-74.9\%)}}
& (-0.086) & (-45.1\%) \\
\addlinespace[2pt]

\rowcolor{methodbg}
\textbf{AutoL2S-SFT}
& 0.558 & 4886
& 0.798 & 2416
& 0.394 & 3492
& 0.929 & \cellcolor{bestcolor}{\textbf{488}}
& 0.436 & 6459
& 0.133 & 15399
& 0.657 & \cellcolor{secondcolor}{\textbf{1064}} \\
\rowcolor{methodbg}
& (-0.032) & (-34.2\%)
& (-0.026) & (-55.1\%)
& (+0.035) & (-42.3\%)
& (+0.003) & \cellcolor{bestcolor}{\textbf{(-63.1\%)}}
& (-0.008) & (-43.0\%)
& (-0.067) & (-16.8\%)
& (-0.129) & \cellcolor{secondcolor}{\textbf{(-46.5\%)}} \\
\addlinespace[2pt]

\rowcolor{methodbg}
\textbf{AutoL2S}
& 0.573 & \cellcolor{bestcolor}{\textbf{2103}}
& 0.804 & \cellcolor{bestcolor}{\textbf{1405}}
& 0.404 & \cellcolor{secondcolor}{\textbf{2798}}
& 0.923 & \cellcolor{secondcolor}{\textbf{663}}
& 0.435 & \cellcolor{bestcolor}{\textbf{2546}}
& 0.100 & \cellcolor{bestcolor}{\textbf{4146}}
& \cellcolor{secondcolor}{\textbf{0.771}} & \cellcolor{bestcolor}{\textbf{1058}} \\
\rowcolor{methodbg}
& (-0.017) & \cellcolor{bestcolor}{\textbf{(-71.7\%)}}
& (-0.020) & \cellcolor{bestcolor}{\textbf{(-73.9\%)}}
& (+0.045) & \cellcolor{secondcolor}{\textbf{(-53.7\%)}}
& (-0.003) & \cellcolor{secondcolor}{\textbf{(-49.8\%)}}
& (-0.009) & \cellcolor{bestcolor}{\textbf{(-77.5\%)}}
& (-0.100) & \cellcolor{bestcolor}{\textbf{(-77.6\%)}}
& \cellcolor{secondcolor}{\textbf{(-0.015)}} & \cellcolor{bestcolor}{\textbf{(-46.8\%)}} \\
\bottomrule
\end{tabular}
}
\label{tab:main}
\end{table*}

\section{Experiments}
\label{sec:experiments}

In this section, we evaluate \Algnameabbr{} as a length-aware reasoning method and, more importantly, analyze how long-short rollout design influences the trade-off between reasoning length, accuracy, and training stability. In addition, we conduct experiments to evaluate the performance of \Algnameabbr{} framework. We aim to answer the following three research questions:
\textbf{RQ1}: How does \Algnameabbr{} perform on LLM reasoning tasks in terms of accuracy and efficiency, when long-short reasoning is explicitly modeled?
\textbf{RQ2}: How do supervision-induced reasoning rollouts affect efficiency-accuracy trade-offs during fine-tuning?
\textbf{RQ3}: What mechanisms govern long-short rollout enable \Algnameabbr{} to reliably preserve performance under compression?

\subsection{Datasets and Baselines}

\noindent\textbf{Datasets} We train the \Algnameabbr{} framework on the Bespoke-Stratos-17k dataset~\cite{bespoke_stratos} and evaluate it on six reasoning benchmarks: Math500~\cite{hendrycks2021measuring}, GPQA-Diamond (GPQA)~\cite{rein2024gpqa}, GSM8K~\cite{cobbe2021gsm8k}, OlympiadBench-Math~\cite{he2024olympiadbench}, AIME24, and MMLU-pro~\cite{wang2024mmlu}. Additional dataset statistics and preprocessing details are provided in Appendix~\ref{apdx:dataset}.
\noindent\textbf{Baseline Methods} We compare \Algnameabbr{} framework with the five state-of-the-art baselines to assess the effectiveness of length reduction and performance preservation. The baselines are listed as follows: R1-Distilled reasoning LLMs (Bespoke-Stratos-3B/7B)~\cite{yeo2025demystifying}, O1-pruner~\cite{luo2025o1}, CoT-Valve~\cite{ma2025cot}, DPO~\cite{rafailov2023direct}, TokenSkip~\cite{xia2025tokenskip}, and AlphaOne~\footnote{Not compatible with Llama-family with vLLM.}~\cite{zhang2025alphaone}. More details are listed in Appendix~\ref{apdx:baseline}.

\subsection{Experimental Settings}
\noindent\textbf{Evaluation of Efficient LLM Reasoning.} Following the settings of~\cite{luo2025o1, yeo2025demystifying}, we evaluate reasoning efficiency using two metrics: (1) accuracy and (2) output token length, which directly reflects the amount of computation incurred during inference with efficiency-accuracy trade-off. The goal is to preserve reasoning performance while minimizing token usage, as shorter outputs under autoregressive decoding directly reduce inference computation. Shorter generations correspond to more concise reasoning and lower decoding cost, while longer generations indicate increased reasoning effort.

\vspace{0.1cm}
\noindent\textbf{Implementation Details.}
To demonstrate the flexibility of \Algnameabbr{} across different LLM backbones, we train the framework using two non-reasoning base LLMs: Llama3.2-3B-Instruct~\cite{touvron2023llama} and Qwen2.5-7B-Instruct. The short reasoning samples are generated via rejection sampling with sampling numbers $k \in \mathbb{N}$ using the Qwen2.5-Math-7B-Instruct model, following the settings of~\cite{yeo2025demystifying, yang2025thinking}.
We filter out duplicate question-answer pairs that appear with both \texttt{<EASY>} and \texttt{<Long Trigger>} after rejection sampling, retaining only the pairs associated with \texttt{<EASY>} in such cases. We employ DeepSeek-R1~\cite{guo2025deepseek} as the strong reasoning-capable teacher model for generating $L$ and Qwen2.5-Math-7B-Instruct~\cite{qwen2.5} as a short CoT teacher for generating $S$. More details are in Appendix~\ref{apdx:prompt}.

\subsection{Reasoning Efficiency of \Algnameabbr{} (RQ1)}
We first report the overall accuracy and efficiency of \Algnameabbr{} across multiple reasoning benchmarks. Improvements in efficiency should be interpreted jointly with accuracy preservation and reasoning length. Additional results from repetition experiments are provided in Appendix~\ref{apdx:add_exp}.
We calculate the improvement percentile relative to the Bespoke-Stratos-3B/7B model, a strong baseline finetuned on the Bespoke-Stratos-17k. We conclude the observations as follows:
\begin{itemize}[leftmargin=10pt]
    \itemsep=-.5pt
    \item \textbf{Baseline Comparison.} Table~\ref{tab:main} reports reasoning accuracy and generated token length across benchmarks. \Algnameabbr{} achieves substantial reductions in reasoning length while preserving accuracy compared to strong baselines. Notably, these gains are obtained by explicitly modeling long and short reasoning as distinct rollout modes, rather than relying on implicit or fixed-length rollouts as in prior methods. Compared to baselines, \Algnameabbr{} achieves the best efficiency-accuracy trade-off and further compresses reasoning paths by up to 71.7\% with negligible accuracy degradation, confirming the effectiveness of the proposed length-aware fine-tuning.
    \item \textbf{Rejection Sampling.} We observe that moderate rejection sampling in the SFT stage benefits the rollout generation for RL stage training. We conduct a study denoted as "w/o RJ," where rejection sampling is disabled by setting $k=0$ during the SFT stage, and compare it against the default setting with \Algnameabbr{} with rejection sampling $k=8$. Although applying rejection sampling may lead to a minor accuracy drop after supervised fine-tuning, it substantially improves the quality of the resulting long-short reasoning behavior during training. In particular, the generated reasoning becomes significantly shorter while the accuracy degradation is largely mitigated, yielding relative improvements of $\sim$71\% and $\sim$53\% for the 3B and 7B models, respectively. Overall, rejection sampling improves \Algnameabbr{}’s accuracy-efficiency trade-off.
    \item \textbf{Efficiency Analysis.} Appendix~\ref{apdx:pf} presents the trade-off between accuracy vs. token usage, demonstrating the best efficiency-accuracy trade-offs compared to all other baselines.
\end{itemize}

\begin{figure}[t!]   
\centering
\includegraphics[width=0.49\textwidth]{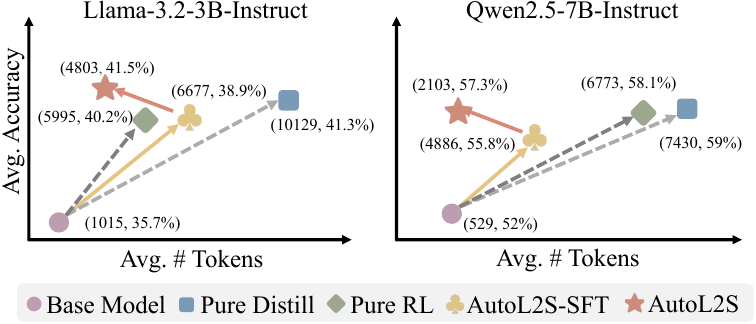}
\vspace{-0.6cm}
\caption{Optimization trajectories of \Algnameabbr{}, showing the trade-off between reasoning length and accuracy.}
\vspace{-.4cm}
\label{fig:trace}
\end{figure}

\begin{table*}[h]
\caption{Comparison of Different Rollout Configurations for Training AutoL2S under Qwen2.5-7B-Instruct}
\vspace{-0.2cm}
\centering
\small
\resizebox{\textwidth}{!}{%
\begin{tabular}{l|cc|cc|cc|cc|cc|cc|cc}
\toprule
\textbf{Method} & \multicolumn{2}{c|}{\textbf{Average}} & \multicolumn{2}{c|}{\textbf{MATH500}} & \multicolumn{2}{c|}{\textbf{GPQA}} & \multicolumn{2}{c|}{\textbf{GSM8K}} & \multicolumn{2}{c|}{\textbf{Olympiad}} & \multicolumn{2}{c|}{\textbf{AIME24}} & \multicolumn{2}{c}{\textbf{MMLU-pro}} \\
\cmidrule(lr){2-3} 
\cmidrule(lr){4-5} 
\cmidrule(lr){6-7} 
\cmidrule(lr){8-9} 
\cmidrule(lr){10-11}
\cmidrule(lr){12-13}
\cmidrule(lr){14-15}
& Acc & Len & Acc & Len & Acc & Len & Acc & Len & Acc & Len & Acc & Len & Acc & Len \\
\midrule
AutoL2S-SFT & 0.558 & 4886 & 0.798 & 2416 & 0.394 & 3492 & 0.929 & 488 & 0.436 & 6459 & 0.133 & 15399 & 0.657 & 1064 \\
AutoL2S & 0.573 & 2103 & 0.804 & 1405 & 0.404 & 2798 & 0.923 & 663 & 0.435 & 2546 & 0.100 & 4146 & 0.771 & 1058 \\
\midrule
w/ Force-short & 0.547 & 2399 & 0.778 & 1305 & 0.399 & 2544 & 0.924 & 578 & 0.438 & 2598 & 0.100 & 5898 & 0.643 & 1468 \\
\quad \textit{\color{gray} $\Delta$ vs AutoL2S} & \color{gray}-0.026 & \color{gray}+14.1\% & \color{gray}-0.026 & \color{gray}-7.1\% & \color{gray}-0.005 & \color{gray}-9.1\% & \color{gray}+0.001 & \color{gray}-12.8\% & \color{gray}+0.003 & \color{gray}+2.0\% & \color{gray}+0.000 & \color{gray}+42.3\% & \color{gray}-0.128 & \color{gray}+38.8\% \\
\addlinespace[0.1cm]
w/ Force-long & 0.570 & 2428 & 0.796 & 1606 & 0.369 & 2529 & 0.923 & 620 & 0.453 & 2726 & 0.167 & 5695 & 0.714 & 1389 \\
\quad \textit{\color{gray} $\Delta$ vs AutoL2S} & \color{gray}-0.003 & \color{gray}+15.4\% & \color{gray}-0.008 & \color{gray}+14.3\% & \color{gray}-0.035 & \color{gray}-9.6\% & \color{gray}+0.000 & \color{gray}-6.5\% & \color{gray}+0.018 & \color{gray}+7.1\% & \color{gray}+0.067 & \color{gray}+37.4\% & \color{gray}-0.057 & \color{gray}+31.3\% \\
\bottomrule
\end{tabular}
}
\label{tab:easy}
\end{table*}

% \begin{table*}[h]
% \caption{Ablation studies of different annotation strategies on AutoL2S-SFT.}
% \vspace{-0.2cm}
% \centering
% \small
% % \resizebox{\textwidth}{!}{%
% \begin{tabular}{l|cc|cc|cc|cc|cc}
% \toprule
% \textbf{Method} & \multicolumn{2}{c|}{\textbf{Average}} 
%                 & \multicolumn{2}{c|}{\textbf{MATH500}} 
%                 & \multicolumn{2}{c|}{\textbf{GPQA}} 
%                 & \multicolumn{2}{c|}{\textbf{GSM8K}} 
%                 & \multicolumn{2}{c}{\textbf{Olympiad}} \\
% \cmidrule(lr){2-3} 
% \cmidrule(lr){4-5} 
% \cmidrule(lr){6-7} 
% \cmidrule(lr){8-9} 
% \cmidrule(lr){10-11}
% & Acc & Len & Acc & Len & Acc & Len & Acc & Len & Acc & Len \\
% \midrule
% Long-only Distill      & 0.638 & 6019 & 0.824 & 5383 & 0.359 & 6049 & 0.926 & 1321 & 0.444 & 11322 \\
% Long-short Separated Distill        &   0.626    &   4250   & 0.798 & 3123 & 0.389 & 5096 & 0.919 & 864  & 0.399 & 7919 \\
% Short-Long Distill      &    0.610   &   1798   & 0.758 & 3239 & 0.414 & 1754 & 0.883 & 701  & 0.384 & 1499 \\
% \midrule\midrule
% Long-Short Distill (RJ $k=0$) & 0.660 & 4512 & 0.800 & 3468 & 0.434 & 4777 & 0.934 & 735  & 0.470 & 9068 \\
% Long-Short Distill (RJ $k=8$) & 0.639 & 3214 & 0.798 & 2416 & 0.394 & 3492 & 0.929 & 488 & 0.436 & 6459 \\
% \bottomrule
% \end{tabular}
% % }
% \label{tab:l2s}
% \end{table*}

\begin{table}[t]
\centering
\small
\caption{Performance of different annotation format.}
\vspace{-0.2cm}
\begin{tabular}{lcc}
\toprule
\textbf{Method} & \textbf{Avg. Acc} & \textbf{Avg. Len} \\
\midrule
Long-only Distill            & 0.668 & 5213 \\
Long-short Separated Distill & 0.644 & 3682 \\
Short-Long Distill           & 0.622 & 2059 \\
\midrule\midrule
Long-Short Distill (w/o RJ)         & 0.674 & 3910 \\
Long-Short Distill   & 0.643 & 2784 \\
\bottomrule
\end{tabular}
\vspace{-0.2cm}
\label{tab:l2s}
\end{table}

\subsection{Impact of Training Dynamics (RQ2)}
\label{apdx:trace}

In this section, we analyze the training dynamics of \Algnameabbr{} to better understand how the training paradigm affects the trade-off between reasoning efficiency and accuracy.
Figure~\ref{fig:trace} illustrates the optimization trajectories of \Algnameabbr{} on both the 3B and 7 B-based models, showing the trade-off between reasoning length (measured by the average number of generated tokens) and accuracy. Compared to the base instruction model and long-only distillation, \Algnameabbr{} leverages paired long-short distillation to achieve both higher accuracy and shorter reasoning paths. Compared to Pure RL, \Algnameabbr{} leverages adaptive long-short rollouts for training and achieves a better trade-off between accuracy and efficiency. Overall, these trajectories show that enabling instance-wise long-short reasoning allows \Algnameabbr{} to substantially reduce reasoning length while preserving task performance.

\subsection{Impact of Annotation Format (RQ2)}
\label{sec:exp-long-annotation-impact}

We next analyze how the annotation format influences the effective long-short rollout behavior learned during training. Rather than directly controlling rollout composition, we reinterpret long-short reasoning annotation as a proxy that modulates the optimization signal for short-mode learning. By varying the availability and quality of short-mode supervision, we implicitly alter how frequently and reliably the model adopts short rollouts at inference time. To further disentangle these effects, we additionally report accuracy and reasoning length conditioned on rollout mode (long vs. short), revealing distinct performance profiles across modes. (1) \textit{Long-only Distill} represents the original distillation from only long reasoning in the Bespoke-Stratos-17k reasoning dataset; (2) \textit{Short-long Distill} switches the position of long and short reasoning path; and (3) \textit{Long-short Separated Distill} separately constructs the long and short CoT reasoning paths. All results are demonstrated in Table~\ref{tab:l2s}. Compared with other formats of long-short term annotation, we observe that \textit{Long-Short Distill} achieves the best performance in terms of accuracy preservation and output length.

\begin{figure*}[t!]
    \centering
    \includegraphics[width=1\linewidth, trim=0 0 0 0, clip]{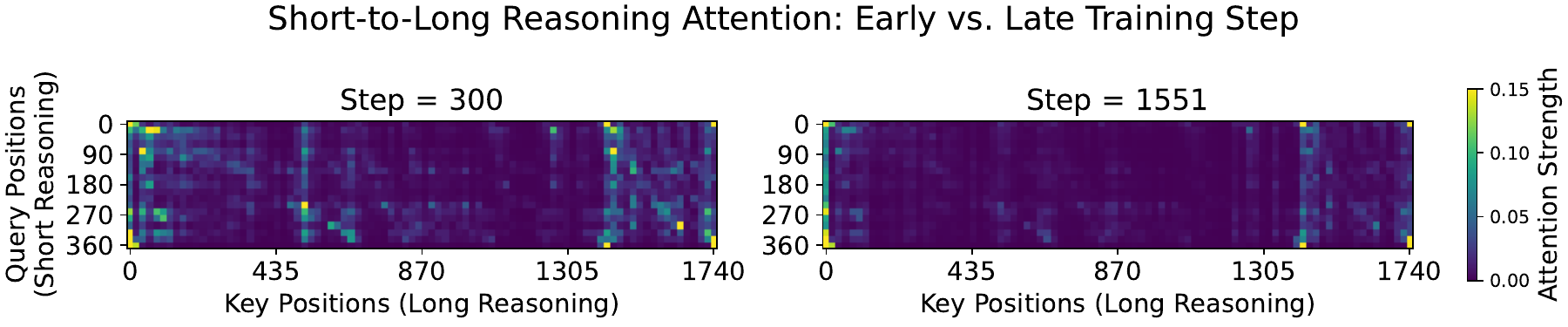}
    \vspace{-0.5cm}
    \caption{Comparison of attention maps at early and late training steps of \Algnameabbr{}. Step 1551 corresponds to the final training step. Given the long sequence lengths, we group every 20 tokens together to calculate attention scores between long and short reasoning paths for better visualization.}
    \label{fig:attn_comparison}    
\end{figure*}

\subsection{Adaptive Rollout Behaviors (RQ2)}
\label{sec:exp-force}
In this section, we analyze the reliability of the <EASY> token as a rollout selector, examining when the model chooses short reasoning and whether such decisions preserve correctness that benefits training \Algnameabbr{}.
We conduct the ablation studies: (1) ``w/ Force-Short'' refers to the setting where \texttt{<Short Trigger>} is always used to generate only short rollout, and (2) ``w/ Force-Long'' denotes the setting where \texttt{<Longer Trigger>} is consistently used to initiate long-only rollout generation. The results are showcased in Table~\ref{tab:easy}.
Compared to the ``Force-Long'' case, \Algnameabbr{} obtains a similar reasoning accuracy on average while generating around 15\% shorter reasoning lengths. Furthermore, we compare \Algnameabbr{} with ``w/ Force-Short'' variants. We observe that \Algnameabbr{} outperforms the ``w/ Force-Short'' in both reasoning accuracy and length. An explanation may be that uniformly enforcing shorter reasoning introduces noisy training signals: aimless truncating reasoning paths can remove necessary intermediate steps, leading to substantial accuracy degradation. In contrast, \Algnameabbr{} instead reinforces reasoning behaviors with adaptive rollouts, allowing the model to learn when shorter reasoning is appropriate. This enables more reliable trade-offs between accuracy and efficiency during training.

\subsection{Impact of Rejection Sampling (RQ2)}
\begin{figure}[t!]   
\centering
\includegraphics[width=0.495\textwidth]{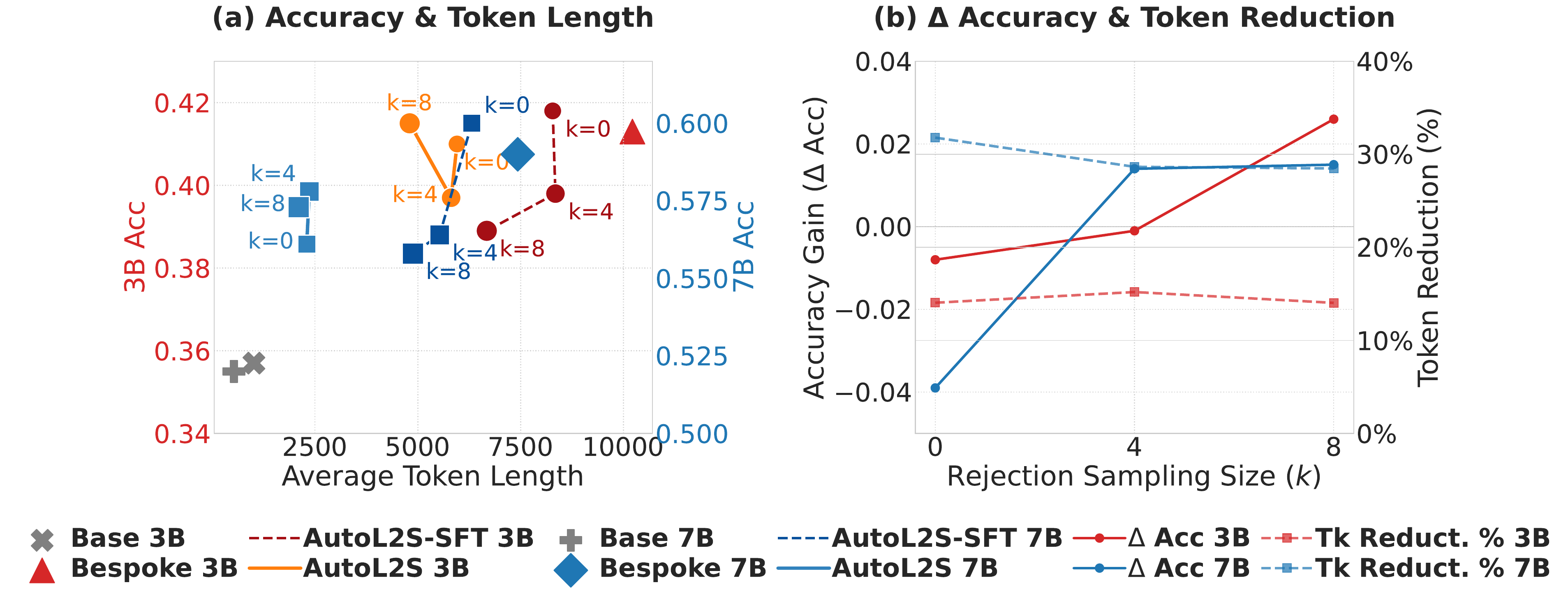}
\vspace{-0.6cm}
\caption{Accuracy and Efficiency Trade-off on different rejection sampling sizes $k$. (a) The Trade-off between Accuracy and Efficiency. (b) Performance Improvement from \Algnameabbr{}-SFT to \Algnameabbr{}}
\vspace{-.4cm}
\label{fig:rj}
\end{figure}

In this section, we conduct a sensitivity analysis on the rejection size $k$. Specifically, we evaluate $k\in\{0, 4, 8\}$ under both the Llama-3.2-3B-Instruct and Qwen2.5-7B-Instruct models. The results are reported in Figure~\ref{fig:rj}. Recall that the larger size of rejection sampling $k$ would lead to a larger proportion of short CoT paths appearing in training data $\mathcal{D}$. In Figure~\ref{fig:rj}-(a), we observe that using a larger size $k$ leads to a more favorable trade-off between reasoning accuracy and generation length under \Algnameabbr{}. This trend is consistent across both base models, with $k=8$ achieving the best overall trade-off compared to $k=0$ and $k=4$. Additionally, a larger size $k$ results in shorter reasoning lengths on average in the SFT stage, but causes more accuracy degradation. This may occur because minimizing reasoning length increases the information bottleneck in supervision, reducing redundancy that can otherwise stabilize learning for complex reasoning cases.
In Figure~\ref{fig:rj}-(b), the results reveal that $k=8$ leads to the best improved rate of accuracy-efficiency trade-offs compared to $k=0$ and $k=4$ for the SFT model. One possible reason is that a larger $k$ encourages in SFT supervision tends to favor aggressively shorter reasoning paths, resulting in shorter but less robust reasoning, and consequently, lower accuracy with shorter reasoning lengths when used alone. When such SFT models (i.e., larger $k$) are used as reference policies for RL fine-tuning, outcome-based feedback restores accuracy while retaining the efficiency-oriented behavior, resulting in improved accuracy-efficiency trade-offs with even shorter reasoning lengths. Thus, we select $k=8$ as our final configuration for \Algnameabbr{} in Table~\ref{tab:main}. 

\subsection{Mechanism behind the Auto Long-short Reasoning (RQ3)}
In this section, we discuss the mechanism explanation of \Algnameabbr{} training. To assess the mechanism behind, Figure~\ref{fig:attn_comparison} presents the attention map comparisons across different training steps of \Algnameabbr{}, highlighting the benefit of the concatenation order used in \textit{Long-Short Distill}. In the early stages of training (i.e., Figure~\ref{fig:attn_comparison} left side: training step 300), we observe that long CoT reasoning paths significantly impact the attention patterns of short CoT reasoning paths, indicating that long-form reasoning benefits the learning of short reasoning generation. As training progresses till the end (i.e., Figure~\ref{fig:attn_comparison} right side: training step 1551), the correlation between long and short CoT reasoning paths significantly diminishes, indicating that they evolve into two distinct components. This separation explains why \Algnameabbr{} is effective and flexible in switching to easy questions simply using the \texttt{<Short Trigger>} when the \texttt{<EASY>} token is presented during inference. The phenomenon again meets the properties of Lemma~\ref{lemma:1}, where long CoT reasoning paths provide auxiliary information for short-path learning. This also explains the reason why the direct use of \texttt{<Short Trigger>} remains effective, without introducing dummy key-value pairs or modifying positional encodings.

\section{Conclusion}
This paper presents Auto Long-Short Reasoning (\Algnameabbr{}), a distillation framework that mitigates overthinking while preserving accuracy. By pairing verified long and short CoT paths, \Algnameabbr{} learns a lightweight switching token that enables instance-wise selection between concise and extended reasoning, and further refines efficiency through post-distillation GRPO-style training without enforcing rigid length constraints.
Experiments show that \Algnameabbr{} reduces reasoning length by up to 71\% with minimal performance loss, achieving favorable trade-offs between token usage, inference time, and accuracy. These results demonstrate the effectiveness of structured long-short supervision, combined with the proposed \Algnameabbr{} framework, for efficient reasoning in distilled LLMs.

\section*{Limitations}

While AutoL2S effectively balances reasoning accuracy and efficiency, its performance depends on the availability and quality of paired long and short reasoning annotations, which may be costly to obtain for some domains. In addition, the binary distinction between long and short reasoning paths may not fully capture more fine-grained variations in reasoning complexity. Our efficiency measurements focus on autoregressive decoding settings, and the gains may differ under alternative inference paradigms. We leave extensions to richer reasoning taxonomies and broader deployment settings to future work.

\bibliography{custom}

\clearpage
\appendix

\onecolumn
\section*{Appendix}
\label{sec:appendix}

\section{Related Work}
\paragraph{Reasoning-capable LLMs.} Recent advancements in LLMs have significantly enhanced their reasoning capabilities, exemplified by large reasoning models such as OpenAI o1~\cite{openai_learning_to_reason} and DeepSeek-R1~\cite{guo2025deepseek}, and QwQ-32B~\cite{qwen_qwq_32b_preview}. OpenAI o1~\cite{openai_learning_to_reason} introduces advanced reasoning mechanisms designed to tackle complex problems, such as mathematical and programming tasks. Similarly, DeepSeek-R1~\cite{guo2025deepseek} enhances reasoning abilities by employing RL to incentivize effective reasoning behaviors. Additionally, DeepSeek-R1 curates specialized reasoning datasets, enabling the explicit distillation of reasoning capabilities into smaller models through SFT. 

\paragraph{Efficient LLM Reasoning.}
% RL//Knowledg distialltion SFT//Tokenskip
Thinking steps of LLMs have become longer, leading to the ``overthinking problem''~\cite{chen2024not, sui2025stop}. To mitigate lengthy responses and reasoning processes, several works have been conducted to shorten the thinking steps and produce more concise reasoning~\cite{sui2025stop}.
\textit{RL-based methods} aim to encourage full-length reasoning models to generate concise thinking steps or train non-reasoning models to learn efficient reasoning by incorporating a length-aware reward~\cite{team2025kimi, luo2025o1, aggarwal2025l1, yeo2025demystifying, shen2025dast, hou2025thinkprune}. Specifically, they propose designing a length-based score to penalize excessively lengthy responses, complementing original rewards (e.g., format reward and accuracy reward).
Kimi K1.5~\cite{team2025kimi} calculates a length reward based on the response length relative to the shortest and longest responses. 
L1~\cite{aggarwal2025l1} modifies the training data with the designated
length constraint instruction, and then add the length reward. 
O1-Pruner~\cite{luo2025o1} introduces the length-harmonizing reward, which calculates the ratio of lengths between the reference model and predicted model along with the accuracy-based constraints.

\textit{SFT-based methods} curate variable-length CoT training datasets to fine-tune overthinking reasoning models for shorter reasoning paths or to equip non-reasoning models with efficient reasoning capabilities~\cite{han2024token, xia2025tokenskip, ma2025cot, yu2025z1efficienttesttimescaling, cui2025stepwise}. Specifically, based on long CoT reasoning paths, they curate shorter yet accurate CoT reasoning paths as training data.
Token-skip~\cite{xia2025tokenskip} leverages LLMLingua~\cite{jiang2023llmlingua} to compress lengthy CoT responses into shorter ones based on semantic scores, and then fine-tunes the model for efficient reasoning.
CoT-Valve~\cite{ma2025cot} controls the magnitude of LoRA~\cite{hu2022lora} weights to generate variable-length CoT training data, which are then used to fine-tune an efficient reasoning model.
Token-Budget~\cite{han2024token} assigns specific token budgets to prompts in order to generate shorter reasoning steps, and these concise CoT examples are then used for model fine-tuning.

\section{Details of Evaluation Dataset}
\label{apdx:dataset}
We train the \Algnameabbr{} framework under the Bespoke-Stratos-17k~\cite{bespoke_stratos} dataset and assess the framework on the long-to-short reasoning task under four different reasoning datasets. The details of the assessment datasets are provided as follows:
\begin{itemize}[leftmargin=*]
    \item \textbf{Math500}~\cite{hendrycks2021measuring}: A challenging benchmark consisting of 500 high-quality math word problems that require multi-step symbolic reasoning.
    \item \textbf{GPQA-Diamond (GPQA)}~\cite{rein2024gpqa}: The Graduate-Level Physics Question Answering (GPQA) dataset contains 198 multiple-choice questions from graduate-level physics exams.
    \item \textbf{GSM8K}~\cite{cobbe2021gsm8k}: A widely-used benchmark comprising 1319 grade school-level math word problems.
    \item \textbf{Olympiad Bench Math (Olympiad)}~\cite{he2024olympiadbench}: A collection of 674 math competition problems inspired by middle and high school mathematics Olympiad competitions. 
    \item \textbf{AIME}~\cite{aime24}: A benchmark consisting of 30 problems from the 2024 American Invitational Mathematics Examination. 
    \item \textbf{MMLU-Pro}~\cite{wang2024mmlu}: A robust multi-task benchmark that enhances evaluation rigor by expanding answer options and curating high-quality problems across diverse disciplines to mitigate random guessing.

\end{itemize}

\section{Details of Baseline Implementation}
\label{apdx:baseline}

\subsection{Bespoke-Stratos}
We implement this baseline by fully fine-tuning language models on the Bespoke-Stratos-17k dataset, which comprises 17,000 examples of questions, long-form reasoning traces, and corresponding answers. The resulting model serves as an oracle reference for reasoning performance.

Following standard SFT procedures, training is performed by minimizing the standard cross-entropy loss over the input sequence. We employ the AdamW optimizer with a learning rate of $1\mathrm{e}{-5}$ and a batch size of 32. Fine-tuning is conducted for three epochs on two NVIDIA A100 80GB GPUs with mixed-precision training enabled. For the 7B base model, we directly utilize the publicly released checkpoint \href{https://huggingface.co/VanWang/Bespoke-Stratos-7B-repro-SFT}{VanWang/Bespoke-Stratos-7B-repro-SFT}.

\subsection{O1-pruner}
O1-pruner introduces a Length-Harmonizing Reward, integrated with a GRPO-style loss, to optimize the policy model $\pi_\theta$ and reduce the length of generated chain-of-thought (CoT) reasoning. Considering the effectiveness of off-policy training with pre-collected data, O1-pruner adopts an off-policy training approach by sampling from the reference model $\pi_{\mathrm{SFT}}$ rather than from $\pi_\theta$. Specifically, the training procedure consists of two steps: (1) generating CoT samples using $\pi_{\mathrm{SFT}}$, and (2) fine-tuning the policy model with the proposed GRPO-style objective based on the generated samples.

In our implementation, we follow the original experimental setting and reproduce the method based on its official repository.\footnote{\url{https://github.com/StarDewXXX/O1-Pruner}} For training, we sample 5,000 problems from the Bespoke-Stratos-17k dataset and generate 16 solutions for each problem. We then perform length-harmonizing fine-tuning for one epoch to jointly optimize both output length and answer correctness. To ensure fair comparison with our method, we use Bespoke-Stratos-3B/7B as the reference model and set the maximum sequence length to 10,240 tokens when training.

\subsection{CoT-Valve}

COT-Valve is designed to enable models to generate reasoning chains of varying lengths. It controls the length of reasoning by linearly combining the LoRA weights of the distilled long-form reasoning CoT and the non-reasoning model. For the specific Long to Short CoT task, it has three stages: (1) finetune the LLM base model on a long-cot dataset using Lora to identify a direction in the parameter space that control the length of generated CoT(2) merge Lora weights with the base model at varying interpolation ratios generate models and use them construct datasets containing CoT of decreasing lengths (3) finetuning the distilled reasoning model with the generated dataset in a progressive way, where the model is trained with shorter reasoning path samples between epochs. This progressive training strategy enables the model to gradually compress its reasoning while maintaining correctness. 

In our implementation, we follow the original configuration in CoT-Valve. The LoRA rank and LoRA alpha are set to 32 and 64, respectively, for both the first and third stages.
In the first stage, we finetune the non-reasoning models Llama-3.2-3B-Instruct/Qwen2.5-7B-Instruct on the Bespoke-Stratos-17k dataset for three epochs using Lora. The learning rate is 4e-5 and the batch size is 64. 
In the second stage, we apply LoRA weight interpolation with coefficients 0.8 and 0.6. Due to resource constraints, we randomly sample 2,000 questions for each interpolated model to generate responses, and retain only those samples with correct answers.
In the third stage, the model obtained in the first stage is further fine-tuned for 2 epochs on each type of generated dataset, using the same learning rate of 4e-5 and a batch size of 64.

\section{Theorem Statements and Proofs}
\label{apdx:thm}
In this section, we present and prove Lemma~\ref{lemma:1} and Theorem~\ref{apex:thm:2}, 
with accompanying remarks to provide intuitive explanations.

\setcounter{lemma}{0}
\setcounter{theorem}{0}

\vspace{0.1cm}
\begin{lemma}[\textbf{Concatenation Advantage for Long--Short CoT Training}]
Let $X$ denote the input, $L=(\ell_1,\dots,\ell_{T_L})$ the long-CoT token sequence, 
and $S=(s_1,\dots,s_{T_S})$ the short-CoT token sequence, 
with training order $L$ to $S$. Then, the conditional entropy $H(\cdot|\cdot)$ of the next short token satisfies:
\begin{equation}
    H(S_t \mid X,L,S_{<t}) \;\le\; H(S_t \mid X,S_{<t}), 
\quad \forall\, t \in [1, T_S].
\end{equation}

Equivalently, averaging across all positions with the improvement quantified as
\begin{equation}
\frac{1}{T_S}\sum_{t=1}^{T_S} 
\Big[ H(S_t \mid X,S_{<t}) - H(S_t \mid X,L,S_{<t}) \Big]
= \frac{1}{T_S}\sum_{t=1}^{T_S} I(S_t;L \mid X,S_{<t}) \;\ge 0.
\end{equation}
Thus, the long CoT reasoning path $L$ provides additional mutual information $I(\cdot|\cdot)$ that strictly increases the entropy of the short CoT reasoning path $S$ whenever $L$ is informative about $S$.
\label{lemma:1}
\end{lemma}

\begin{proof}
The inequality follows directly from the fact that conditioning reduces entropy: 
adding $L$ to the conditioning set cannot increase the uncertainty of $S_t$. 
Formally, for each $t\in[T_S]$,
\[
H(S_t \mid X,L,S_{<t}) \;\le\; H(S_t \mid X,S_{<t}).
\]
Averaging over $t$ yields the stated inequality.  

The gap between the two sides can be expressed as the conditional mutual information:
\[
\frac{1}{T_S}\sum_{t=1}^{T_S} \big[ H(S_t \mid X,S_{<t}) - H(S_t \mid X,L,S_{<t}) \big]
= \frac{1}{T_S}\sum_{t=1}^{T_S} I(S_t;L \mid X,S_{<t}) \;\ge 0.
\]

In the realizable training case under long CoT reasoning path distillation, the model is optimized with 
the per-token cross-entropy objective
\[
\mathrm{CE}(S \mid \mathcal{C})
= \frac{1}{T_S}\sum_{t=1}^{T_S} 
\mathbb{E}\big[-\log p_\theta(s_t \mid \mathcal{C},S_{<t})\big],
\]
where the context $\mathcal{C}$ is either $(X)$ or $(X,L)$.  
When $p_\theta$ matches the true distribution, the cross-entropy coincides with the entropy above.  
Thus, the same inequality carries over to cross-entropy:
\[
\mathrm{CE}(S \mid X,L) \;\le\; \mathrm{CE}(S \mid X),
\]
with the gap equal to the average conditional mutual information.  
Finally, since perplexity is defined as 
$\mathrm{PPL}(S \mid \mathcal{C})=\exp(\mathrm{CE}(S \mid \mathcal{C}))$, 
the inequality extends directly to perplexity:
\[
\mathrm{PPL}(S \mid X,L) \;\le\; \mathrm{PPL}(S \mid X).
\]
\end{proof}

\begin{theorem}[\textbf{Rollout Adaptation with \texttt{<EASY>} Token}]
Let $p_\theta^{L}(\cdot \mid x)$ and $p_\theta^{S}(\cdot \mid x)$ denote the predictive distributions when decoding with the long and short CoT reasoning paths $L=(\ell_1,\dots,\ell_{T_L})$ and $S=(s_1,\dots,s_{T_S})$, respectively. Given an input $x \in \mathcal{Y}$, define the per-instance risks as
\begin{align}
    J_{S}(x) &= 
    \mathbb{E}\!\left[ D\!\Big(p_\theta^{S}(\cdot\mid x)\,\Big\|\,p_\theta^{L}(\cdot\mid x)\Big)\right]
    \;+\; \lambda \,\mathbb{E}[T_{S}(x)], \\
    J_{L}(x) &= 
    \lambda \,\Big(\mathbb{E}[T_{L}(x)] + c_\pi\Big),
\end{align}
where $D(\cdot\|\cdot)$ is a statistical divergence,  
$T_{S}(x)$ and $T_{L}(x)$ denote the token lengths of the short and long CoT reasoning paths,  
$\lambda > 0$ is the per-token cost, and $c_\pi \ge 0$ is a fixed overhead for invoking the long path. Then the optimal adaptation policy is
\begin{equation}
\pi^*(x) =
\begin{cases}
0 & \text{if } J_{S}(x) < J_{L}(x) \quad (\text{choose short}), \\
1 & \text{otherwise} \quad (\text{choose long}).
\end{cases}
\end{equation}
\label{apex:thm:2}
\end{theorem}

\begin{proof} We provide the proof within the following six steps.

\medskip
\noindent
\textbf{Assumptions from AutoL2S Design.}
Let $\mathcal{Y}$ be the input space with data distribution $\mathcal{D}$.
Assume $D(\cdot\|\cdot)\ge 0$ is a statistical divergence for which 
$\mathbb{E}[D(p_\theta^{S}(\cdot\mid x)\,\|\,p_\theta^{L}(\cdot\mid x))]$ exists,
and the token lengths $T_S(x),T_L(x)$ are nonnegative random variables with finite expectations.
Let an adaptation policy be a measurable mapping 
$\pi:\mathcal{Y}\to\{0,1\}$, where $\pi(x){=}0$ chooses short reasoning CoT and $\pi(x){=}1$ chooses long reasoning CoT. For a policy $\pi$, define the population risk
\[
\mathcal{R}(\pi)
:= \mathbb{E}_{x\sim\mathcal{D}}
\Big[\,J_S(x)\,\mathbf{1}\{\pi(x)=0\} + J_L(x)\,\mathbf{1}\{\pi(x)=1\}\,\Big].
\]
By the assumptions above, $\mathcal{R}(\pi)$ is well-defined and finite.

\medskip
\noindent
\textbf{Step 1 (Reduction to deterministic policies).}
Consider any \emph{randomized} policy that, for a fixed $x$, chooses short with
probability $\alpha(x)\in[0,1]$ and long with probability $1-\alpha(x)$.
Its \emph{conditional} (on $x$) contribution to risk equals
\[
\alpha(x)\,J_S(x) + (1-\alpha(x))\,J_L(x)
= J_L(x) + \alpha(x)\,\Delta(x),
\quad \text{where }\;\Delta(x):=J_S(x)-J_L(x).
\]
Since this expression is linear in $\alpha(x)$, its minimum over $\alpha(x)\in[0,1]$
is always achieved at an extreme point $\alpha(x)\in\{0,1\}$:
\[
\alpha^\star(x)=
\begin{cases}
1, & \text{if }\Delta(x)<0,\\
0, & \text{if }\Delta(x)>0,\\
\text{any in }[0,1], & \text{if }\Delta(x)=0.
\end{cases}
\]
Hence, randomization cannot improve over a deterministic rule, and it suffices to prove that it optimizes over a deterministic policy $\pi$.

\medskip
\noindent
\textbf{Step 2 (Pointwise decomposition).}
For any deterministic $\pi$,
\[
\mathcal{R}(\pi)
= \mathbb{E}\big[J_L(x)\big] + \mathbb{E}\big[\Delta(x)\,\mathbf{1}\{\pi(x)=0\}\big].
\]
The first term does not depend on $\pi$, so minimizing $\mathcal{R}(\pi)$ reduces to minimizing
the second term. Because the expectation is taken with respect to $\mathcal{D}$ and
the integrand depends on $\pi$ only through the indicator, this is a pointwise decision:

\medskip
\noindent
\textbf{Step 3 (Pointwise optimal action).}
For a fixed $x$:
\[
\min_{a\in\{0,1\}} \big\{\,\Delta(x)\,\mathbf{1}\{a=0\}\,\big\}
=
\begin{cases}
\Delta(x), & \text{if } a=0 \text{ and } \Delta(x)<0,\\
0,         & \text{if } a=1 \text{ or } \Delta(x)\ge 0,
\end{cases}
\]
which is achieved by choosing $a{=}0$ (short) when $\Delta(x)<0$, and $a{=}1$ (long) otherwise.
Thus, the Bayes-optimal policy is
\[
\pi^\star(x)=
\begin{cases}
0, & \text{if } \Delta(x)<0\;\; \big(\text{i.e., } J_S(x)<J_L(x)\big),\\
1, & \text{otherwise}.
\end{cases}
\]
This is exactly the threshold rule stated in the theorem.

\medskip
\noindent
\textbf{Step 4 (Existence and uniqueness).}
Existence follows because the pointwise minimum is always attained by an action in $\{0,1\}$.
Uniqueness holds everywhere except on the \emph{tie set}
$\{x:\Delta(x)=0\}$ where both actions yield the same risk; changing $\pi^\star$ on this set does not alter $\mathcal{R}(\pi^\star)$.
Hence, the optimal policy is unique almost surely (up to ties).

\medskip
\noindent
\textbf{Step 5 (Explicit threshold and interpretation).}
Expanding $\Delta(x)$ gives
\[
\Delta(x)
= \underbrace{\mathbb{E}\!\left[D\Big(p_\theta^{S}(\cdot\mid x)\,\Big\|\,p_\theta^{L}(\cdot\mid x)\Big)\right]}_{\text{predictive distribution divergence}}
\;+\; \lambda\,\Big(\mathbb{E}[T_S(x)] - \mathbb{E}[T_L(x)] - c_\pi\Big).
\]
Thus $\pi^\star(x){=}0$ (choose short) iff the divergence penalty is outweighed by the token savings:
\[
\mathbb{E}\!\left[D\Big(p_\theta^{S}\,\Big\|\,p_\theta^{L}\Big)\right]
\;<\; \lambda\,\Big(\mathbb{E}[T_L(x)] + c_\pi - \mathbb{E}[T_S(x)]\Big).
\]
Equivalently, \emph{choose short when predicted distributions are sufficiently close \emph{and} the token savings are large enough}.

\medskip
\noindent
\textbf{Step 6 (Comparative statics).}
The decision boundary moves monotonically: increasing $c_\pi$ or the long/short length gap $\mathbb{E}[T_L]-\mathbb{E}[T_S]$ makes short more favorable; increasing the divergence or decreasing the length gap makes long more favorable. Increasing $\lambda$ amplifies the weight on token savings, thus favoring short when $\mathbb{E}[T_L]+c_\pi>\mathbb{E}[T_S]$.
\end{proof}

\begin{remark}
\textrm{Theorem~\ref{apex:thm:2} establishes that an optimal adaptation strategy 
between long and short CoT reasoning paths always exists and is essentially 
unique, reducing to a deterministic threshold rule.  
The policy selects the short path whenever the predictive distribution of the short rationale is sufficiently close to that of the long reasoning 
while offering enough token savings to offset the overhead of using the long path.  
This shows that the \texttt{<EASY>} token is not an ad hoc mechanism, 
but corresponds to a Bayes-optimal decision that balances semantic fidelity 
and inference efficiency.  
Together with Lemma~\ref{lemma:1}, this highlights that the long reasoning paths 
not only improve the learnability of the short reasoning paths during training, 
but also guide optimal switching at inference time.}
\end{remark}

\newpage
\section{Details of Implementation and Instruction Prompt and Triggers}
\label{apdx:prompt}
In this section, we introduce the format of instruction prompts and triggers that we utilized in our \Algnameabbr{} framework.
\subsection{Details of Implementation Settings}
All experiments for the 7B base model are conducted using four NVIDIA A100 80G GPUs, while those for the 3B base model utilize two NVIDIA A100 80GB GPUs. We leverage the Transformers library for fine-tuning and vLLM for efficient inference. Fine-tuning is performed using the AdamW optimizer with a learning rate of 1e-5. The temperature is fixed at 0.7 in both \Algnameabbr{} and baselines, ensuring that the output reasoning sequences are fully generated without truncation. 

\subsection{System prompt and Inference Prompt}
\begin{tcolorbox}[title=System Prompt / Inference Prompt, colback=gray!5, colframe=blue!30!black, boxrule=0.5pt, arc=1mm, left=2mm, right=2mm, top=1mm, bottom=1mm, fonttitle=\bfseries,sidebyside align=top seam,]
Your role as an assistant is to solve problems using one of two ways.\\
First, think step-by-step and give the answer. The solution should be formatted as follows:\\
\texttt{<|begin\_of\_solution|>} \\
\texttt{\{precise and concise solution\}} \\
\texttt{<|end\_of\_solution|>} \\

Second, your output thoroughly explores questions through a systematic, long-term thinking process before providing the final, precise, and accurate solutions. This requires engaging in a comprehensive cycle of analysis, summarizing, exploration, reassessment, reflection, backtracing, and iteration to develop a well-considered thinking process.\\

Please structure your response into two main sections: Thought and Solution.\\

In the Thought section, detail your reasoning process using the specified format:\\
\texttt{<|begin\_of\_thought|>} \texttt{\{thought with steps separated with '\textbackslash n\textbackslash n'\}} \texttt{<|end\_of\_thought|>}\\

Each step should include detailed considerations such as analyzing questions, summarizing relevant findings, brainstorming new ideas, verifying the accuracy of the current steps, refining any errors, and revisiting previous steps.\\

In the Solution section, based on various attempts, explorations, and reflections from the Thought section, systematically present the final solution that you deem correct. The solution should remain a logical, accurate, concise expression style and detail the necessary steps needed to reach the conclusion, formatted as follows:\\
\texttt{<|begin\_of\_solution|>} \\
\texttt{\{final formatted, precise, and clear solution\}} \\
\texttt{<|end\_of\_solution|>}\\

Now, try to solve the following question through the above guidelines:
\end{tcolorbox}

\subsection{Special Token and Long-short Trigger}

\begin{tcolorbox}[title=Special Easy Token and Long-short Trigger, colback=gray!5, colframe=blue!30!black, boxrule=0.5pt, arc=1mm, left=2mm, right=2mm, top=1mm, bottom=1mm, fonttitle=\bfseries,sidebyside align=top seam,]
\texttt{<EASY>} Token: \texttt{<specialLong>}\\

\texttt{Short Trigger}: This is a trigger to ensure the model’s upcoming output \texttt{<EASY>}. \\ 

\textbf{Long Trigger}: Let’s consider this problem in a \texttt{<pureLong>} way.

\end{tcolorbox}

\newpage

\section{Additional Experimental Results of \Algnameabbr{}}

\label{apdx:add_exp}

\subsection{\textcolor{black}{Pareto Front of Reasoning Accuracy and Efficiency}}
\label{apdx:pf}
\begin{figure}[h!]
\centering
\begin{subfigure}[t]{0.46\textwidth}
    \includegraphics[width=\textwidth]{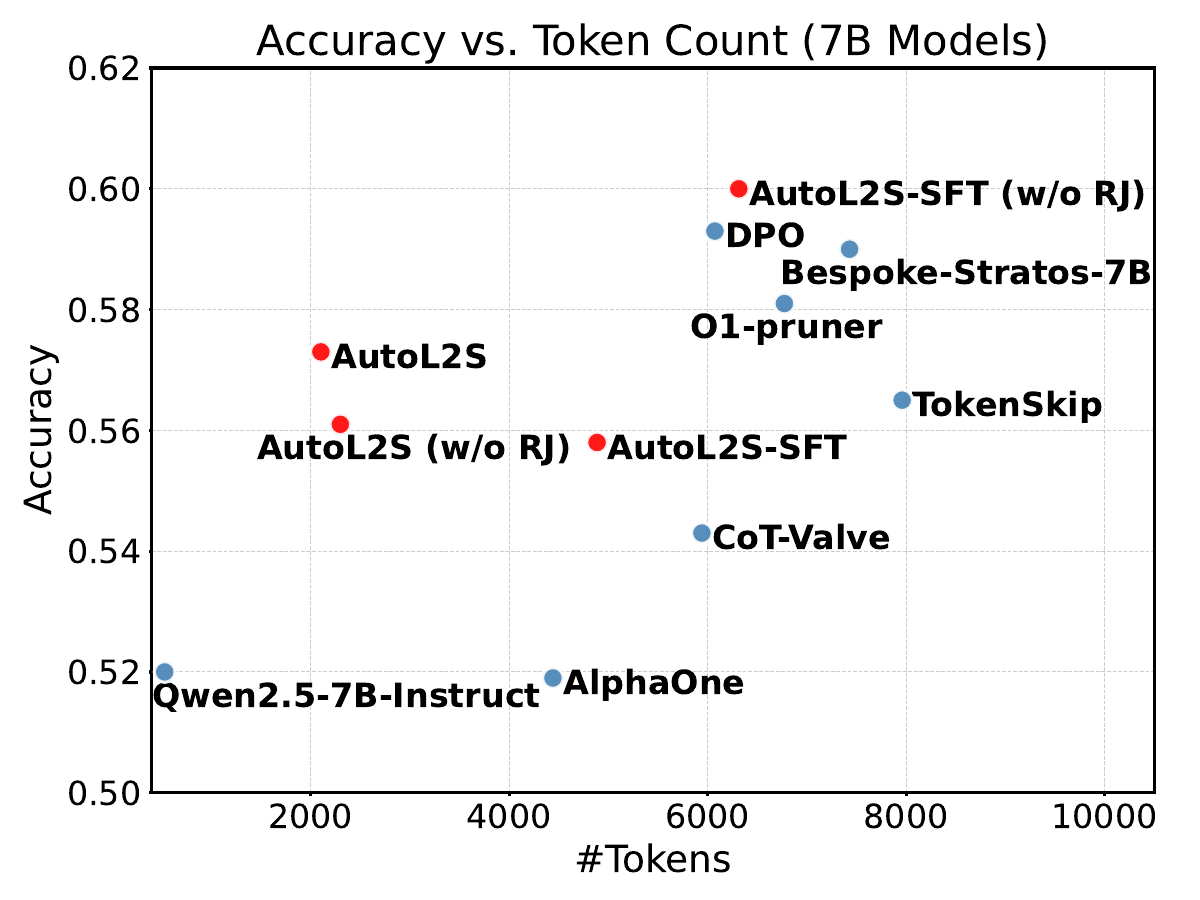}
    \caption{Accuracy vs.\ Token Count (7B Models)}
\end{subfigure}
\hfill
\begin{subfigure}[t]{0.46\textwidth}
    \includegraphics[width=\textwidth]{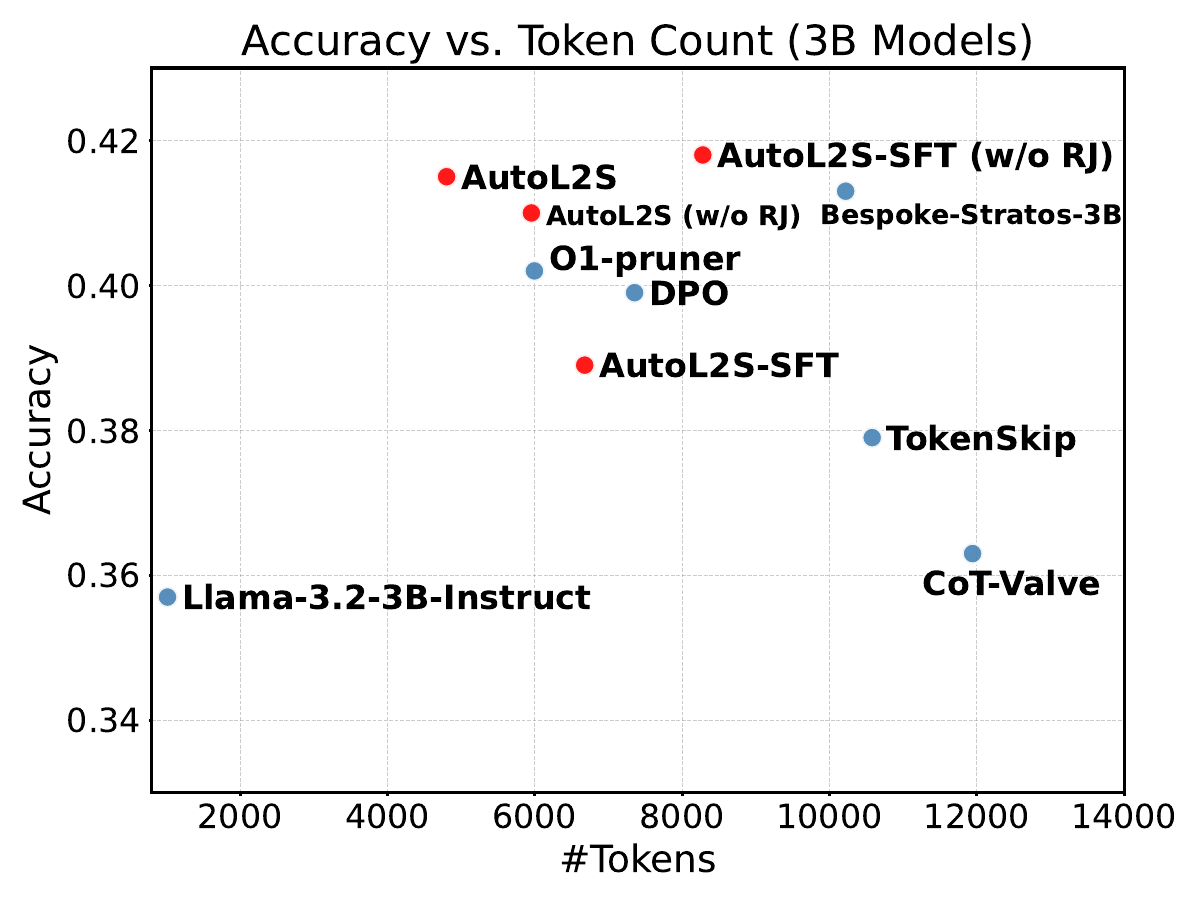}
    \caption{Accuracy vs.\ Token Count (3B Models)}
\end{subfigure}
\caption{\label{fig:pf}
\textcolor{black}{Efficiency-accuracy trade-off of AutoL2S. AutoL2S variants consistently improve accuracy under substantially lower inference cost.}}
\label{apdx:fig:pf}
\end{figure}

In this section, we showcase the Pareto Front of different methods in terms of their accuracy and token counts. The results are shown in Figure~\ref{apdx:fig:pf}. We observe that \Algnameabbr{} obtains the best trade-off between accuracy and reasoning efficiency.

\subsection{Robustness Analytics of \Algnameabbr{}-SFT}
\label{apdx:experiments}

To assess the robustness of our method, we further evaluated AutoL2S on both 3B and 7B models under three different runs with different random seeds. The reported values correspond to the mean and standard deviation with the same settings presented in Section~\ref{sec:experiments}. The \textbf{bold} numbers represent the best performance, and \underline{underline} refers to the second best among the settings.

Based on the average performance, \Algnameabbr{}-SFT outperforms CoT-Valve by achieving higher accuracy and generating shorter reasoning paths. Compared to O1-pruner, \Algnameabbr{}-SFT produces shorter reasoning paths while maintaining comparable average accuracy across all four reasoning benchmarks. Furthermore, \Algnameabbr{}-SFT achieves nearly the same average accuracy as the oracle SFT R1-distilled models (i.e., Bespoke-Stratos-3B/7B), while significantly reducing reasoning path length. This presents the same observation showcased in Section~\ref{sec:experiments}.

Considering standard deviation, \Algnameabbr{}-SFT continues to outperform both the oracle SFT R1-distilled models and other baselines, offering better accuracy and lower average token usage. For example, with AutoL2S based on Qwen2.5-7B-Instruct, the performance remains the best among all methods, while also achieving the shortest reasoning lengths. These results demonstrate that \Algnameabbr{}-SFT has both competitive and robust performance in efficient reasoning tasks.

\begin{table*}[h]
\centering
\caption{Evaluation results of AutoL2S based on Qwen2.5-3B-Instruct.(mean ± std)}
\resizebox{\textwidth}{!}{
\begin{tabular}{lcccccccccc}
\toprule
\textbf{} & \multicolumn{2}{c}{\textbf{Average}} & \multicolumn{2}{c}{\textbf{MATH500}} & \multicolumn{2}{c}{\textbf{GPQA}} & \multicolumn{2}{c}{\textbf{GSM8K}} & \multicolumn{2}{c}{\textbf{Olympiad}} \\
\cmidrule(lr){2-3} \cmidrule(lr){4-5} \cmidrule(lr){6-7} \cmidrule(lr){8-9} \cmidrule(lr){10-11}
& Acc & Len & Acc & Len & Acc & Len & Acc & Len & Acc & Len \\
\midrule
Qwen2.5-3B-Instruct & 0.479 & 777 & 0.622 & 806 & 0.349 & 770 & 0.679 & 376 & 0.266 & 1158 \\
Bespoke-Stratos & 0.516 & 8931 & 0.636 & 9246 & 0.308 & 10129 & \underline{0.848} & 1624 & 0.272 & 14724 \\
CoT-Valve & 0.484 & 5889 & 0.602 & 4980 & 0.258 & 6898 & 0.805 & 1660 & 0.270 & 10017 \\
O1-pruner & \textbf{0.535} & 6686 & \textbf{0.704} & 6769 & 0.283 & 7348 & \textbf{0.859} & 1210 & \textbf{0.295} & 11416 \\
\midrule
AutoL2S-SFT & \makecell{0.523\\$\pm$0.006} & \makecell{5083\\$\pm$737} & \makecell{0.656\\$\pm$0.015} & \makecell{4287\\$\pm$605} & \makecell{\underline{0.322}\\\underline{$\pm$0.003}} & \makecell{\textbf{4018}\\$\boldsymbol{\pm}$\textbf{941}} & \makecell{0.830\\$\pm$0.026} & \makecell{1109\\$\pm$224} & \makecell{\underline{0.284}\\\underline{$\pm$0.023}} & \makecell{10919\\$\pm$1293} \\
AutoL2S-SFT(w/ RJ $k=4$) & \makecell{\underline{0.524}\\ \underline{$\pm$0.009}} & \makecell{\underline{3569}\\\underline{$\pm$506}} & \makecell{0.646\\$\pm$0.016} & \makecell{\underline{2713}\\\underline{$\pm$135}} & \makecell{\textbf{0.347}\\$\boldsymbol{\pm}$\textbf{0.015}} & \makecell{\underline{4118}\\\underline{$\pm$514}} & \makecell{0.826\\$\pm$0.003} & \makecell{\underline{503}\\\underline{$\pm$4}} & \makecell{0.278\\$\pm$0.007} & \makecell{\underline{6942}\\\underline{$\pm$1915}} \\
AutoL2S-SFT(w/ RJ $k=8$) & \makecell{0.523\\$\pm$0.007} & 
\makecell{\textbf{3255}\\$\boldsymbol{\pm}$\textbf{548}} & \makecell{\underline{0.671}\\ \underline{$\pm$0.021}} & \makecell{\textbf{2523}\\$\boldsymbol{\pm}$\textbf{200}} & \makecell{0.317\\$\pm$0.008} & \makecell{4135\\$\pm$598} & \makecell{0.825\\$\pm$0.004} & \makecell{\textbf{417}\\$\boldsymbol{\pm}$\textbf{41}} & \makecell{0.280\\$\pm$0.005} & \makecell{\textbf{5947}\\$\boldsymbol{\pm}$\textbf{1796}} \\

\bottomrule
\end{tabular}
}
\end{table*}

\begin{table*}[h]
\centering
\caption{Evaluation results of AutoL2S based on Qwen2.5-7B-Instruct.(mean ± std)}
\resizebox{\textwidth}{!}{
\begin{tabular}{lcccccccccc}
\toprule
\textbf{} & \multicolumn{2}{c}{\textbf{Average}} & \multicolumn{2}{c}{\textbf{MATH500}} & \multicolumn{2}{c}{\textbf{GPQA}} & \multicolumn{2}{c}{\textbf{GSM8K}} & \multicolumn{2}{c}{\textbf{Olympiad}} \\
\cmidrule(lr){2-3} \cmidrule(lr){4-5} \cmidrule(lr){6-7} \cmidrule(lr){8-9} \cmidrule(lr){10-11}
& Acc & Len & Acc & Len & Acc & Len & Acc & Len & Acc & Len \\
\midrule
Qwen2.5-7B-Instruct & 0.586 & 435 & 0.748 & 556 & 0.308 & 27 & 0.902 & 260 & 0.384 & 896 \\
Bespoke-Stratos & 0.638 & 6019 & \underline{0.824} & 5383 & 0.359 & 6049 & \underline{0.926} & 1321 & \underline{0.444} & 11322 \\
CoT-Valve & 0.594 & 4747 & 0.730 & 4483 & 0.369 & 4930 & 0.898 & 928 & 0.378 & 8647 \\
O1-pruner & \underline{0.650} & 5267 & \textbf{0.832} & 5104 & \underline{0.399} & 5312 & \textbf{0.936} & 1065 & 0.433 & 9586 \\
\midrule
AutoL2S-SFT & \makecell{\textbf{0.652}\\$\boldsymbol{\pm}$\textbf{0.007}} & \makecell{4348\\$\pm$306} & \makecell{0.795\\$\pm$0.005} & \makecell{3278\\$\pm$240} & \makecell{\textbf{0.431}\\$\boldsymbol{\pm}$\textbf{0.006}} & \makecell{4590\\$\pm$532} & \makecell{0.923\\$\pm$0.011} & \makecell{595\\$\pm$150} & \makecell{\textbf{0.460}\\$\boldsymbol{\pm}$\textbf{0.010}} & \makecell{8932\\$\pm$335} \\
AutoL2S-SFT(w/ RJ $k=4$) & \makecell{0.630\\$\pm$0.011} & \makecell{\underline{3233}\\\underline{$\pm$474}} & \makecell{0.788\\$\pm$0.017} & \makecell{\underline{2200}\\\underline{$\pm$354}} & \makecell{0.375\\$\pm$0.033} & \makecell{\underline{3103}\\\underline{$\pm$494}} & \makecell{0.915\\$\pm$0.003} & \makecell{\underline{439}\\\underline{$\pm$68}} & \makecell{0.442\\$\pm$0.009} & \makecell{\underline{7190}\\\underline{$\pm$994}} \\
AutoL2S-SFT(w/ RJ $k=8$) & \makecell{0.626\\$\pm$0.013} & \makecell{\textbf{2746}\\$\boldsymbol{\pm}$\textbf{496}} & \makecell{0.785\\$\pm$0.012} & \makecell{\textbf{2019}\\$\boldsymbol{\pm}$\textbf{368}} & \makecell{0.380\\$\pm$0.019} & \makecell{\textbf{2587}\\$\boldsymbol{\pm}$\textbf{799}} & \makecell{0.915\\$\pm$0.015} & \makecell{\textbf{415}\\$\boldsymbol{\pm}$\textbf{75}} & \makecell{0.422\\$\pm$0.016} & \makecell{\textbf{5964}\\$\boldsymbol{\pm}$\textbf{921}} \\
\bottomrule
\end{tabular}
}
\end{table*}

% \vspace{5em}

\clearpage

\subsection{\Algnameabbr{} with Length Penalty Reward}

This experiment evaluates the effect of introducing an explicit length-penalty reward in \Algnameabbr{}. While the penalty further reduces reasoning length beyond the default setting, it also increases the risk of accuracy degradation by encouraging overly aggressive compression. These results highlight the trade-off between enforcing conciseness through explicit rewards and preserving reasoning correctness.

\begin{table*}[h]
\small
\centering
\caption{Effect of length penalty on the performance of 3B models ($r_j=8$).}
\resizebox{\textwidth}{!}{%
\begin{tabular}{l 
c c 
c c 
c c 
c c 
c c 
c c 
c c}
\toprule
\textbf{Method}
& \multicolumn{2}{c|}{\textbf{Average}}
& \multicolumn{2}{c|}{\textbf{MATH500}}
& \multicolumn{2}{c|}{\textbf{GPQA}}
& \multicolumn{2}{c|}{\textbf{GSM8K}}
& \multicolumn{2}{c|}{\textbf{Olympiad}}
& \multicolumn{2}{c|}{\textbf{AIME}}
& \multicolumn{2}{c}{\textbf{MMLU-Pro}} \\
\cmidrule(lr){2-3}
\cmidrule(lr){4-5}
\cmidrule(lr){6-7}
\cmidrule(lr){8-9}
\cmidrule(lr){10-11}
\cmidrule(lr){12-13}
\cmidrule(lr){14-15}
& Acc & Len
& Acc & Len
& Acc & Len
& Acc & Len
& Acc & Len
& Acc & Len
& Acc & Len\\
\midrule

% ===================== 3B Models =====================
% \multicolumn{15}{c}{\textit{Llama-3.2-3B-Instruct}} \\

Llama-3.2-3B-Instruct
& 0.357 & 1015
& 0.404 & 740
& 0.293 & 498
& 0.729 & 203
& 0.147 & 2117
& 0.067 & 2053
& 0.500 & 477 \\
\addlinespace[2pt]

Bespoke-Stratos-3B
& 0.413 & 10219
& 0.574 & 10148
& 0.273 & 8888
& 0.822 & 1387
& 0.246 & 15635
& 0.033 & 21341
& 0.529 & 3912 \\
\addlinespace[2pt]

% % === AutoL2S (rj=0) ===
% AutoL2S (rj=0)
% & 0.418 & 8280
% & 0.552 & 5990
% & 0.389 & 7520
% & 0.823 & 1166
% & 0.206 & 12941
% & 0.067 & 19158
% & 0.471 & 2906 \\
% \addlinespace[2pt]

% % === AutoL2S-Plus (rj=0) w/o len penalty ===
% AutoL2S-Plus (rj=0) 
% & 0.410 & 5954
% & 0.564 & 6508
% & 0.359 & 4569
% & 0.815 & 964
% & 0.230 & 9269
% & 0.033 & 12494
% & 0.457 & 1919 \\
% \addlinespace[2pt]

% % === AutoL2S-Plus (rj=0) ===
% AutoL2S-Plus (rj=0) w/ len
% & 0.410 & 3273
% & 0.534 & 3645
% & 0.273 & 2920
% & 0.829 & 650
% & 0.230 & 4832
% & 0.067 & 6288
% & 0.529 & 1301 \\
% \addlinespace[2pt]

% === AutoL2S (rj=8) ===
AutoL2S-SFT
& 0.389 & 6677
& 0.546 & 4181
& 0.369 & 6165
& 0.800 & 1021
& 0.218 & 10706
& 0.000 & 14775
& 0.400 & 3211 \\
\addlinespace[2pt]

% === AutoL2S-Plus (rj=8) w/o len penalty ===
AutoL2S 
& 0.415 & 4803
& 0.520 & 4116
& 0.273 & 3679
& 0.826 & 819
& 0.193 & 7199
& 0.167 & 11538
& 0.514 & 1469 \\
\addlinespace[2pt]

% === AutoL2S-Plus (rj=8) ===
AutoL2S (w/ len)
& 0.398 & 2190
& 0.550 & 1819
& 0.273 & 2048
& 0.810 & 353
& 0.233 & 3099
& 0.067 & 4971
& 0.457 & 848 \\
% \addlinespace[2pt]

\bottomrule
\end{tabular}
}
\label{tab:len_penalty}
\end{table*}

\section{Full Experimental Results of \Algnameabbr{} on Rejection Sampling Ablation Studies}

This section presents the full experimental results of \Algnameabbr{} under varying rejection sampling sizes. We analyze how different values of $k$ affect reasoning accuracy and generation length during supervised distillation and subsequent refinement. The results highlight the role of rejection sampling in shaping the accuracy-efficiency trade-off of \Algnameabbr{} across models and training stages.

\begin{table*}[h]
\small
\caption{Accuracy (Acc) and Token Length (Len) for 3B and 7B models with different rejection sampling ratios across reasoning benchmarks.}
\centering
\resizebox{\textwidth}{!}{%
\begin{tabular}{l
c c
c c
c c
c c
c c
c c
c c}
\toprule
\textbf{Method}
& \multicolumn{2}{c|}{\textbf{Average}}
& \multicolumn{2}{c|}{\textbf{MATH500}}
& \multicolumn{2}{c|}{\textbf{GPQA}}
& \multicolumn{2}{c|}{\textbf{GSM8K}}
& \multicolumn{2}{c|}{\textbf{Olympiad}}
& \multicolumn{2}{c|}{\textbf{AIME}}
& \multicolumn{2}{c}{\textbf{MMLU-Pro}} \\
\cmidrule(lr){2-3}
\cmidrule(lr){4-5}
\cmidrule(lr){6-7}
\cmidrule(lr){8-9}
\cmidrule(lr){10-11}
\cmidrule(lr){12-13}
\cmidrule(lr){14-15}
& Acc & Len
& Acc & Len
& Acc & Len
& Acc & Len
& Acc & Len
& Acc & Len
& Acc & Len\\
\midrule

% ===================== 3B Models =====================
\multicolumn{15}{c}{\textit{Llama-3.2-3B-Instruct}} \\
\midrule

Llama-3.2-3B-Instruct
& 0.357 & 1015
& 0.404 & 740
& 0.293 & 498
& 0.729 & 203
& 0.147 & 2117
& 0.067 & 2053
& 0.500 & 477 \\
\addlinespace[2pt]

Bespoke-Stratos-3B
& 0.413 & 10219
& 0.574 & 10148
& 0.273 & 8888
& 0.822 & 1387
& 0.246 & 15635
& 0.033 & 21341
& 0.529 & 3912 \\
\addlinespace[2pt]

AutoL2S (w/o RJ)
& 0.418 & 8280
& 0.552 & 5990
& 0.389 & 7520
& 0.823 & 1166
& 0.206 & 12941
& 0.067 & 19158
& 0.471 & 2906 \\
\addlinespace[2pt]

AutoL2S-Plus (w/o RJ)
& 0.410 & 5954
& 0.564 & 6508
& 0.359 & 4569
& 0.815 & 964
& 0.230 & 9269
& 0.033 & 12494
& 0.457 & 1919 \\
\addlinespace[2pt]

AutoL2S (w/ RJ $k=4$)
& 0.398 & 8347
& 0.574 & 5666
& 0.283 & 7546
& 0.812 & 1322
& 0.226 & 12185
& 0.067 & 18134
& 0.429 & 5227 \\
\addlinespace[2pt]

AutoL2S-Plus (w/ RJ $k=4$)
& 0.397 & 5810
& 0.562 & 5517
& 0.288 & 4510
& 0.815 & 876
& 0.230 & 8257
& 0.000 & 12669
& 0.486 & 3029 \\
\addlinespace[2pt]

AutoL2S (w/ RJ $k=8$)
& 0.389 & 6677
& 0.546 & 4181
& 0.369 & 6165
& 0.800 & 1021
& 0.218 & 10706
& 0.000 & 14775
& 0.400 & 3211 \\
\addlinespace[2pt]

AutoL2S-Plus (w/ RJ $k=8$)
& 0.415 & 4803
& 0.520 & 4116
& 0.273 & 3679
& 0.826 & 819
& 0.193 & 7199
& 0.167 & 11538
& 0.514 & 1469 \\
\addlinespace[2pt]

\midrule\midrule
\multicolumn{15}{c}{\textit{Qwen2.5-7B-Instruct}} \\
\midrule

Qwen2.5-7B-Instruct
& 0.520 & 529
& 0.748 & 556
& 0.308 & 27
& 0.902 & 260
& 0.384 & 896
& 0.133 & 1014
& 0.643 & 423 \\
\addlinespace[2pt]

Bespoke-Stratos-7B
& 0.590 & 7430
& 0.824 & 5383
& 0.359 & 6049
& 0.926 & 1321
& 0.444 & 11322
& 0.200 & 18513
& 0.786 & 1989 \\
\addlinespace[2pt]

AutoL2S-SFT (w/o RJ)
& 0.600 & 6314
& 0.800 & 3468
& 0.434 & 4777
& 0.934 & 735
& 0.470 & 9068
& 0.233 & 18332
& 0.729 & 1504 \\
\addlinespace[2pt]

AutoL2S (w/o RJ)
& 0.561 & 2299
& 0.798 & 1601
& 0.414 & 2666
& 0.912 & 707
& 0.439 & 3088
& 0.100 & 4638
& 0.700 & 1091 \\
\addlinespace[2pt]

AutoL2S-SFT (w/ RJ $k=4$)
& 0.564 & 5531
& 0.786 & 2560
& 0.409 & 3495
& 0.917 & 509
& 0.438 & 7991
& 0.133 & 17220
& 0.700 & 1409 \\
\addlinespace[2pt]

AutoL2S (w/ RJ $k=4$)
& 0.578 & 2361
& 0.788 & 1477
& 0.465 & 2707
& 0.921 & 609
& 0.445 & 2450
& 0.133 & 5503
& 0.714 & 1417 \\
\addlinespace[2pt]

AutoL2S-SFT (w/ RJ $k=8$)
& 0.558 & 4886
& 0.798 & 2416
& 0.394 & 3492
& 0.929 & 488
& 0.436 & 6459
& 0.133 & 15399
& 0.657 & 1064 \\
\addlinespace[2pt]

AutoL2S (w/ RJ $k=8$)
& 0.573 & 2103
& 0.804 & 1405
& 0.404 & 2798
& 0.923 & 663
& 0.435 & 2546
& 0.100 & 4146
& 0.771 & 1058 \\
\bottomrule
\end{tabular}
}
\label{tab:rejection_sample_ratio}
\end{table*}

\newpage
\section{Case Studies of \Algnameabbr{}}
\label{apdx:case studies}

In this section, we present additional case studies to demonstrate the efficiency of \Algnameabbr{}. The \Algnameabbr{} responses shown in this section are generated by a model trained on Qwen2.5-7B-Instruct with $rj = 8$. The baseline responses in the first two examples are produced by Bespoke-Stratos-7B~\cite{bespoke_stratos}. We include three types of inference modes: Auto Long Reasoning, Auto Short Reasoning, and AutoL2S without \texttt{<EASY>}.

\subsection{Qualitative Analysis of \Algnameabbr{}}
The \Algnameabbr{} response shown here is generated by a model trained in Qwen2.5-7B-Instruct with $rj = 8$, and the baseline response is produced by Bespoke-Stratos-7B. The case studies demonstrate clear improvements in AutoL2S output after training. Before training, responses were often verbose, meandering, and filled with self-doubt, with frequent backtracking and confusion in applying domain knowledge. This will lead to several rounds of redundant thinking and verification steps. We observe that in some simple questions, complex verification steps can be reduced or even omitted, where LLMs are strong enough to provide the answers. After training with AutoL2S, the outputs became more structured and coherent, with reasoning expressed in a logical step-by-step manner and conclusions stated concisely and decisively. The use of special triggers aligned better with the intended long-short reasoning split, domain knowledge was applied more accurately, and verbose exploration was replaced by tighter yet sufficiently detailed explanations. Overall, training enhanced fluency, confidence, and correctness while reducing redundancy, showing a clear shift from rambling exploration to disciplined, solution-oriented reasoning.

\subsection{Auto Long Reasoning for Regular Questions}
We showcase an example when encountering a regular question. \Algnameabbr{} automatically chooses to reason longer but eliminates the token usage compared to baseline responses.

\begin{tcolorbox}[title=Case Study: The Lollipop Packing Problem, colback=gray!5, colframe=blue!30!black, boxrule=0.5pt, arc=1mm, left=2mm, right=2mm, top=1mm, bottom=1mm, fonttitle=\bfseries,sidebyside align=top seam,]
\textbf{question:} Jean has 30 lollipops. Jean eats 2 of the lollipops. With the remaining lollipops, Jean wants to package 2 lollipops in one bag. How many bags can Jean fill?

\textbf{Baseline Response (445 Tokens):}
\newline
\texttt{<|begin\_of\_thought|>} Okay, let's see. Jean has 30 lollipops. She eats 2 of them. Then she wants to put the remaining ones into bags, with each bag holding 2 lollipops. The question is, how many bags can she fill? Hmm, first step, I need to figure out how many lollipops Jean has left after she eats 2. So originally, she has 30. If she eats 2, that means she subtracts 2 from 30. Let me write that down: 30 minus 2 equals 28. So she has 28 lollipops left. Now, she wants to put 2 lollipops in each bag. So, to find out how many bags she can fill, I need to divide the number of remaining lollipops by 2. That would be ...[251 tokens omitted]

\vspace{0.5em}

\textbf{AutoL2S Response (136 Tokens):}  
\newline
\texttt{<|begin\_of\_solution|>} \textbf{<EASY><Short Trigger>} This is a trigger to ensure the model’s upcoming output \texttt{<EASY>}. Let's approach this step by step. Jean starts with 30 lollipops. She eats 2, so we need to subtract those from the total. 30 minus 2 equals 28. So, Jean has 28 lollipops left. Now, Jean wants to put 2 lollipops in each bag. To find out how many bags she can fill, we divide the number of remaining lollipops by the number of lollipops per bag. 28 divided by 2 equals 14. Therefore, Jean can fill 14 bags. \texttt{<|end\_of\_solution|>}
\end{tcolorbox}

\vspace{-0.2cm}
\begin{tcolorbox}[title=Case Study: Auto Long Reasoning for Regular Questions, colback=gray!5, colframe=blue!30!black, boxrule=0.5pt, arc=1mm, left=2mm, right=2mm, top=1mm, bottom=1mm, fonttitle=\bfseries,sidebyside align=top seam,]
\textbf{question:} Melissa works as a pet groomer. This week, she has 8 dogs that need to be bathed, 5 cats that need their nails clipped, 3 birds that need their wings trimmed, and 12 horses that need to be brushed. If she splits the grooming jobs evenly over the days, how many animals will she groom each day of the week?

\textbf{Baseline Response (2207 Tokens):}
\newline
\texttt{<|begin\_of\_thought|>}

Okay, let's see. Melissa is a pet groomer, and she has different animals to groom this week. The problem is asking how many animals she'll groom each day if she splits the jobs evenly over the days. Alright, so first, I need to figure out the total number of animals she has to groom. Let me list them out:

- Dogs: 8 that need bathing.\\
- Cats: 5 that need their nails clipped.\\
- Birds: 3 that need their wings trimmed.\\
- Horses: 12 that need brushing. \\
...[1889 tokens omitted]...

\texttt{<|end\_of\_thought|>}

\texttt{<|begin\_of\_solution|>}

Melissa has a total of 8 dogs, 5 cats, 3 birds, and 12 horses to groom. To find the total number of animals:

\[
8 \, (\text{dogs}) + 5 \, (\text{cats}) + 3 \, (\text{birds}) + 12 \, (\text{horses}) = 28 \, (\text{total animals})
\]

The problem asks how many animals she will groom each day if the jobs are split evenly over the days. Assuming she works 7 days a week, the calculation is:

\[
\frac{28 \, (\text{total animals})}{7 \, (\text{days})} = 4 \, (\text{animals per day})
\]

Thus, the final answer is \boxed{4}.

\texttt{<|end\_of\_solution|>}

\vspace{0.5em}

\textbf{AutoL2S Response (1351 Tokens):}  
\newline
\texttt{<|begin\_of\_thought|>}

Let’s consider this problem in a \texttt{<pureLong>} way. Okay, let's see. So the problem is about Melissa, who's a pet groomer. She has different animals to groom: 8 dogs, 5 cats, 3 birds, and 12 horses. The question is asking how many animals she will groom each day if she splits the jobs evenly over the days of the week.

First, I need to figure out the total number of grooming jobs she has. 

...[1112 tokens omitted]...

\texttt{<|end\_of\_thought|>}

\texttt{<|begin\_of\_solution|>}

Melissa has a total of 8 dogs, 5 cats, 3 birds, and 12 horses to groom. Adding these together gives:

\[
8 + 5 + 3 + 12 = 28
\]

She needs to split these 28 grooming jobs evenly over 7 days of the week. Dividing the total number of jobs by the number of days gives:

\[
28 \div 7 = 4
\]

Thus, the number of animals she will groom each day is:

\[
\boxed{4}
\]

\texttt{<|end\_of\_solution|>}

\end{tcolorbox}

\subsection{Auto Short Reasoning for EASY Questions}
\Algnameabbr{} can automatically switch to from long to short path without sacrificing the reasoning performance.

\begin{tcolorbox}[title=Case Study: Auto Short Reasoning for EASY Questions, colback=gray!5, colframe=blue!30!black, boxrule=0.5pt, arc=1mm, left=2mm, right=2mm, top=1mm, bottom=1mm, fonttitle=\bfseries,sidebyside align=top seam,]
\textbf{question:} Andrew plans a road trip from New Jersey to Rochester. It takes 6 days to travel by bus and half as many days to travel by car from New Jersey to Rochester. How many days will he travel if he plans to go to Rochester by bus and return from Rochester by car?

\vspace{0.5em}

\textbf{Baseline Answer (573 Tokens):} 
\newline
\texttt{<|begin\_of\_thought|>}

Okay, so Andrew is planning a road trip from New Jersey to Rochester. The problem says it takes 6 days to travel by bus, and half as many days by car. Then he wants to go by bus and return by car. I need to figure out the total number of days he'll travel.

First, let me parse the information. The trip from New Jersey to Rochester by bus takes 6 days. Then, the return trip by car would take half as many days as the bus trip. So, half of 6 days is 3 days. That makes sense because the problem states "half as many days to travel by car," so it's half the bus time.

So, going by bus: 6 days. Coming back by car: 3 days. To find the total travel time, I need to add both legs of the journey. That would be 6 days + 3 days. Let me do the math: 6 + 3 equals 9. So, the total travel time is 9 days.

Wait a second, let me make sure I didn't miss anything. The problem says he plans to go to Rochester by bus and return from Rochester by car. So, it's a round trip. The key here is that the return trip is from Rochester back to where he started, which is New Jersey. So, the distances are the same, but the mode of transportation changes.

The bus trip to Rochester is 6 days, car trip back is half that, so 3 days. Adding them together gives 9 days total travel time. That seems straightforward. Is there any part of this that could be interpreted differently?

For example, maybe the problem is referring to something else, like including time in New Jersey or Rochester? But the problem specifically mentions the travel time by bus and car for the same distance. So, it's just the round trip, with each leg's time as given.

Another check: if the bus takes 6 days, then half of that is indeed 3 days. So, 6 + 3 is 9. Yeah, that seems right. I don't see any other factors here. The answer should be 9 days.

\texttt{<|end\_of\_thought|>}

\vspace{0.5em}

\texttt{<|begin\_of\_solution|>}

Andrew plans to travel from New Jersey to Rochester by bus, which takes 6 days. The return trip by car takes half as many days as the bus trip, which is \(6 \div 2 = 3\) days. The total travel time is the sum of both trips:

\[
6 \text{ days (bus)} + 3 \text{ days (car)} = 9 \text{ days}
\]

The final answer is \boxed{9}.

\texttt{<|end\_of\_solution|>}

\vspace{0.5em}

\textbf{AutoL2S Answer (200 Tokens):}  
\texttt{<|begin\_of\_solution|>}  %\textbf{<EASY><Short Trigger>} 

This is a trigger to ensure the model’s upcoming output \texttt{<EASY>}. To determine the total number of days Andrew will travel, we need to consider both the trip to Rochester and the return trip from Rochester. Let's break it down step by step.

1. **Trip to Rochester by bus:**
   - It takes 6 days to travel by bus from New Jersey to Rochester.

2. **Return trip from Rochester by car:**
   - The problem states that it takes half as many days to travel by car as it does by bus.
   - Since the bus trip takes 6 days, the car trip will take \( \frac{6}{2} = 3 \) days.

3. **Total travel time:**
   - Add the time taken for both trips: \( 6 \text{ days (bus)} + 3 \text{ days (car)} = 9 \text{ days} \).

Therefore, the total number of days Andrew will travel is \( \boxed{9} \).

\texttt{<|end\_of\_solution|>}
\end{tcolorbox}

\subsection{AutoL2S w/o <EASY>}

\begin{tcolorbox}[title=Case Study: AutoL2S w/o <EASY>, colback=gray!5, colframe=blue!30!black, boxrule=0.5pt, arc=1mm, left=2mm, right=2mm, top=1mm, bottom=1mm, fonttitle=\bfseries,sidebyside align=top seam,]
\textbf{question:} Janet’s ducks lay 16 eggs per day. She eats three for breakfast every morning and bakes muffins for her friends every day with four. She sells the remainder at the farmers' market daily for \$2 per fresh duck egg. How much in dollars does she make every day at the farmers' market?

\textbf{AutoL2S w/o <EASY> (986 Tokens):}  

\texttt{<|begin\_of\_thought|>}

<specialLong> We will provide a detailed explanation and solution. Okay, let's see. So Janet's ducks lay 16 eggs every day. That's the starting point. Then she does a couple of things with those eggs. First, she eats three for breakfast every morning, 

...[504 tokens omitted]...

<|end\_of\_thought|>

\texttt{<|begin\_of\_solution|>}

Janet starts with 16 eggs each day. She eats 3 eggs for breakfast and uses 4 eggs to bake muffins, totaling 3 + 4 = 7 eggs used per day. Subtracting this from the daily gives 16 - 7 = 9 eggs left to sell. Each egg is sold for \$2, so the total money made from selling the eggs is 9 × 2 = 18 dollars.

The final answer is \boxed{18}.\texttt{<|end\_of\_solution|>}\texttt{<seperate>}\texttt{<|begin\_of\_solution|>}

This is a trigger to ensure the model’s upcoming output <short>. To determine how much Janet makes every day at the farmers' market, we need to follow these steps:

1. **Calculate the total number of eggs laid by the ducks per day:**
   Janet's ducks lay 16 eggs per day.

2. **Determine the number of eggs Janet eats for breakfast each day:**
   Janet eats 3 eggs for breakfast every morning.

3. **Determine the number of eggs Janet uses to bake muffins each day:**
   Janet bakes muffins for her friends every day with 4 eggs.

4. **Calculate the total number of eggs Janet uses each day:**
   \[
   3 \text{ (eggs for breakfast)} + 4 \text{ (eggs for muffins)} = 7 \text{ eggs}
   \]

5. **Calculate the number of eggs remaining after Janet uses some for breakfast and baking:**
   \[
   16 \text{ (total eggs)} - 7 \text{ (eggs used)} = 9 \text{ eggs}
   \]

6. **Determine the amount of money Janet makes from selling the remaining eggs:**
   Janet sells the remaining 9 eggs at the farmers' market for \$2 per egg.
   \[
   9 \text{ eggs} \times 2 \text{ dollars per egg} = 18 \text{ dollars}
   \]

Therefore, the amount of money Janet makes every day at the farmers' market is \(\boxed{18}\).

\texttt{<|end\_of\_solution|>}

% \vspace{0.5em}

\textbf{AutoL2S w <EASY> (314 Tokens):}  

\texttt{<|begin\_of\_solution|>}  %\textbf{<EASY><Short Trigger>} 

This is a trigger to ensure the model’s upcoming output \texttt{<EASY>}. To determine how much Janet makes every day at the farmers' market, we need to follow these steps:

1. **Calculate the total number of eggs laid by the ducks per day:**
   Janet's ducks lay 16 eggs per day.
...[158 tokens omitted]...

4. **Determine how much Janet makes from selling the remaining eggs:**
   Janet sells the remaining 9 eggs at the farmers' market for \$2 per egg. Therefore, the total amount she makes is:
   \[
   9 \text{ eggs} \times 2 \text{ dollars per egg} = 18 \text{ dollars}
   \]

Thus, the amount Janet makes every day at the farmers' market is \boxed{18} dollars.

\texttt{<|end\_of\_solution|>}
\end{tcolorbox}

\end{document}